\documentclass[lettersize,journal]{IEEEtran}
\usepackage{amsmath,amsfonts}
\usepackage{algorithm}
\usepackage{array}
\usepackage{textcomp}
\usepackage{stfloats}
\usepackage{url}
\usepackage{verbatim}
\usepackage{graphicx}
\usepackage{cite}
\usepackage{CJKutf8}
\usepackage{bbm}
\usepackage{caption}
\usepackage{algpseudocode}
\usepackage{amsmath}
\usepackage{amssymb}
\usepackage{makecell}
\usepackage{xcolor,colortbl}
\usepackage{booktabs}
\usepackage{tabularx} 
\usepackage{makecell} 
\usepackage{subcaption}
\usepackage{graphicx}    
\usepackage{wrapfig}
\usepackage{tcolorbox}  
\tcbuselibrary{listingsutf8}
\definecolor{lightblue}{rgb}{0.93, 0.93, 1.0}
\definecolor{lightpink}{rgb}{1,0.93,0.93}
\definecolor{darkgray}{rgb}{0.66, 0.66, 0.66}
\usepackage{color}
\usepackage{enumitem}
\usepackage{multirow} 

\usepackage{amsthm}
\newtheorem{theorem}{Theorem}
\newtheorem{definition}{Definition}

\setlist{labelindent=\parindent, leftmargin=*}

\begin{document}

\title{Un-mixing Test-time Adaptation under Heterogeneous Data Streams}

\author{
Zixian Su$^*$, Jingwei Guo$^*$, Xi Yang$^\dagger$, Qiufeng Wang, Kaizhu Huang$^\dagger$
\thanks{Zixian Su is with  Beijing Academy of Artificial Intelligence, Beijing, China (E-mail:~zxsu@baai.ac.cn);
Jingwei Guo is with Alibaba Group, Beijing, China (E-mail:~jingweiguo19@outlook.com);
Xi Yang and Qiufeng Wang are  with  Xi'an Jiaotong-Liverpool University, Suzhou, Jiangsu, China (Email:~xi.yang01@xjtlu.edu.cn; qiufeng.wang@xjtlu.edu.cn);
Kaizhu. Huang is with Duke Kunshan University, Kunshan, Jiangsu, China (Email: kaizhu.huang@dukekunshan.edu.cn).}
\thanks{$^*$Equal contribution.}
\thanks{$^\dagger$Corresponding authors.}
}

\maketitle

\begin{abstract}
Deploying deep models in real-world scenarios remains challenging due to significant performance drops under distribution shifts between training and deployment environments. Test-Time Adaptation (TTA) has recently emerged as a promising solution, enabling on-the-fly model adaptation. However, its effectiveness deteriorates in the presence of mixed distribution shifts -- common in practical settings -- where multiple target domains coexist. In this paper, we study TTA under mixed distribution shifts and move beyond conventional whole-batch adaptation paradigms. By revisiting distribution shifts from a spectral perspective, we find that the heterogeneity across latent domains is often pronounced in Fourier space. In particular, high-frequency components encode domain-specific variations, which facilitates clearer separation of samples from different distributions.
Motivated by this observation, we propose to un-mix heterogeneous data streams using high-frequency domain cues, making diverse shift patterns more tractable.
To this end, we propose Frequency-based Decentralized Adaptation (FreDA), a novel framework that decomposes globally heterogeneous data stream into locally homogeneous clusters in the Fourier space. It leverages decentralized learning and augmentation strategies to robustly adapt under mixed domain shifts. 
Extensive experiments across various environments (corrupted, natural, and medical) show the superiority of our method over the state-of-the-arts.
\end{abstract}


\section{Introduction}\label{sec:intro}

\IEEEPARstart{D}{eep} learning models often suffer significant performance degradation when deployed in environments where the data distribution differs from that of the training set -- a challenge known as domain shift~\cite{domainshift2,domainshift}. Recently, Test-Time Adaptation (TTA)~\cite{wang2021tent,chen2022contrastive,wang2022continual,niu2022efficient,niu2022towards,su2024unraveling,press2024rdumb,deyo} has emerged as a promising solution by refining model parameters to better align with the encountered data at inference time. It leverages the incoming data stream for real-time adjustments without the need for retraining on a labeled dataset, enabling swift model adaptation to unpredictable data characteristics during deployment.

Despite their success, current TTA models are often  limited to ideal testing conditions, typically involving homogeneous test samples with similar types of distribution shifts. In reality, distribution shifts are \emph{mixed, overlapping, and even conflicting}~\cite{mt2,mt3,mt4,yao2025scmix}. For instance, {photo management software} handles diverse corruptions, such as noise, blur, compression; medical imaging platforms manage inconsistencies from varied acquisition methods; and autonomous driving systems face fluctuating conditions like lighting, weather, and road types. 

\begin{figure}[t]
    \centering
    \includegraphics[width=0.48\textwidth]{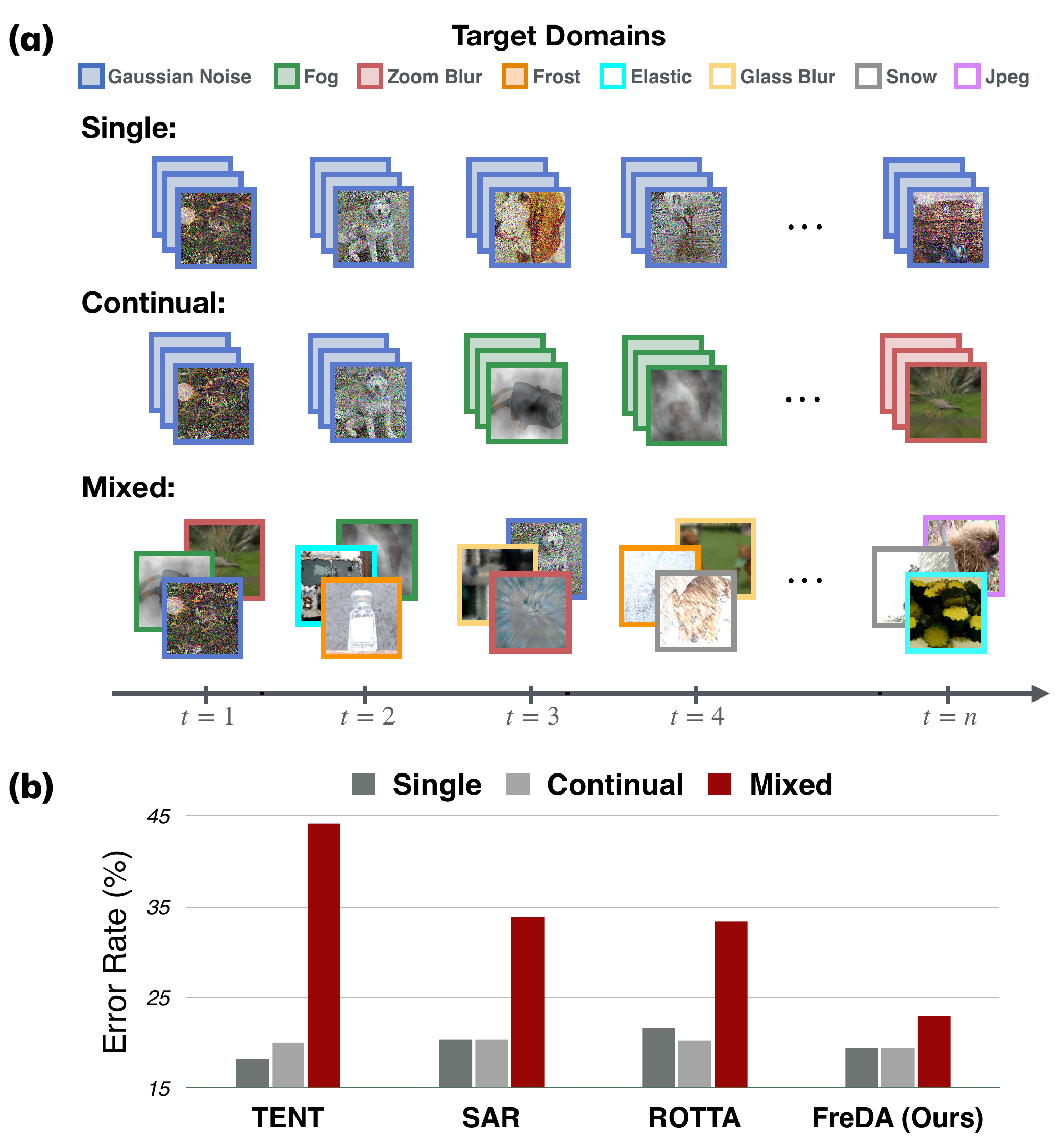}
\caption{
(a) Illustration of different online distribution shifts, ranging from single-domain to continual and mixed-domain shifts, with increasing complexity and realism.
(b) Classification error rate on CIFAR-10-C: while conventional models experience a sharp performance drop as distribution shifts become more complex, our method maintains superior performance, particularly under mixed-domain shifts.
}
\label{fig:tta_problem}
\end{figure}

Such complexity poses serious challenges for existing methods. While recent efforts have extended TTA to continually changing environments~\cite{wang2022continual}, they typically assume a uniform target domain at each time step, as illustrated in the continual setting in Fig.~\ref{fig:tta_problem}~(a). Some methods periodically reset the model to its source pre-trained state~\cite{niu2022towards,press2024rdumb} to mitigate the cumulative effect of sequential distribution shifts on adaptation. Others down-weight outlier target samples~\cite{niu2022efficient,deyo} to suppress the impact of abrupt or rare shifts. Although these strategies aim to improve robustness under dynamic conditions, they lack the capacity to disentangle or localize domain variations within a batch. As a result, they often misalign conflicting domain shifts, failing to be deployed in realistic, heterogeneous data streams (see Fig.~\ref{fig:tta_problem}~(b)).

To address this, we propose shifting from coarse, homogeneous / whole-batch adaptation to a fine-grained strategy that explicitly disentangles heterogeneous shifts. 
Our approach leverages the spectral information to disentangle domain variations, which naturally separates sample features across frequency bands: high-frequency components capture fine details like edges and textures, while low-frequency components represent global structures such as shape and illumination (see Section~\ref{sec:high_freq_effect}). By decomposing inputs accordingly, we are able to better characterize distributional variations and identify diverse shifts. Moreover, the Fourier transform operates directly on raw pixel-level inputs, independent of any pre-trained model, making it robust to large domain gaps and enhancing adaptability in real-world deployments.

\begin{figure*}[t]
    \centering
    \includegraphics[width=0.98\linewidth]{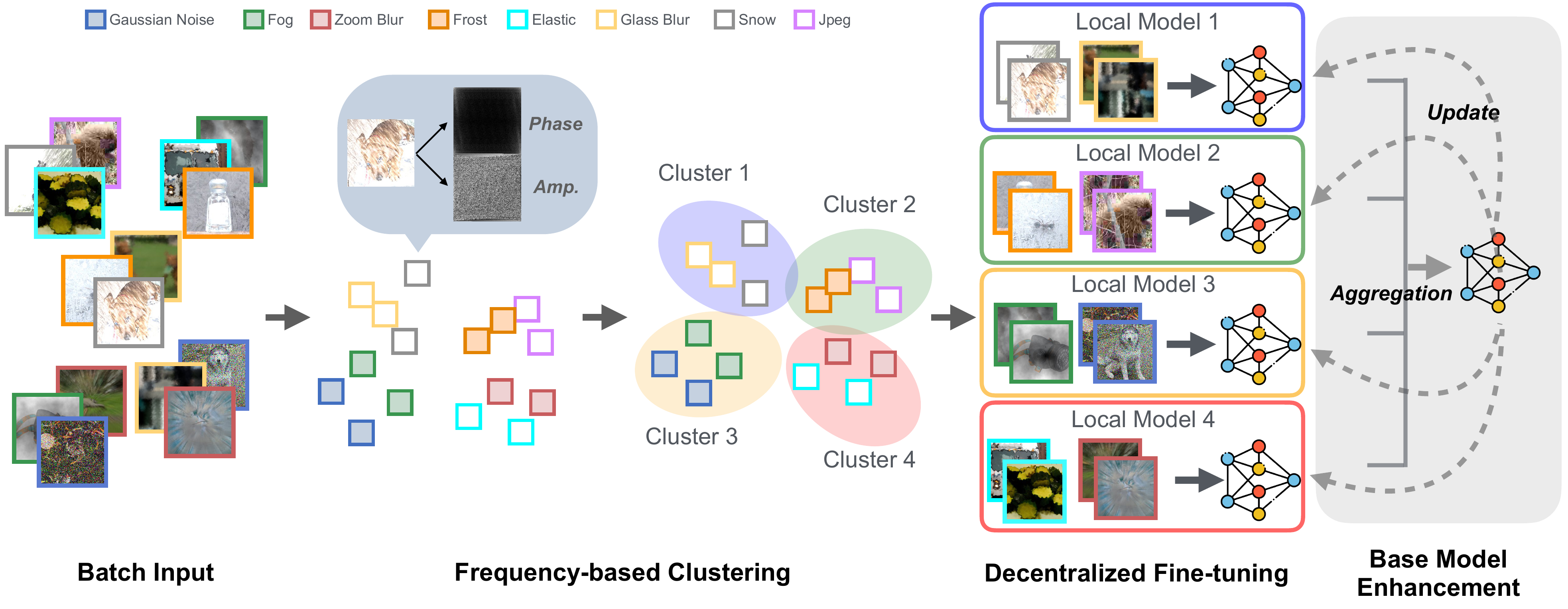}
\caption{Illustration of FreDA framework. 
Heterogeneous data streams are partitioned in Fourier space into locally homogeneous clusters, where local models adapt independently and synchronize weights for intermittent update.
}
\label{fig:tta_pipeline}
\end{figure*}

Building upon this insight, we introduce Frequency-based Decentralized Adaptation (FreDA). FreDA begins by partitioning incoming data in the Fourier space, leveraging high-frequency components to group samples with similar domain characteristics. This transforms globally heterogeneous inputs into locally homogeneous subsets before any adaptation takes place. 
Multiple local models are then deployed, each assigned to a specific cluster, and adapt independently to their respective data streams while periodically uploading their parameters. These intermittent updates help build more stable model weights, reducing potential errors during clustering.
To further enhance robustness, we introduce a Fourier-based augmentation scheme that improves sample quality and reinforces adaptation to shift-specific patterns. To summarize, the main contributions of this work are three-fold:

\begin{itemize} 
    \item  We identify a key limitation of most existing TTA methods -- their neglect of real-world data heterogeneity -- which leads to suboptimal performance when confronted with diverse, mixed distribution shifts.
    \item We propose FreDA, a frequency-based decentralized adaptation framework that leverages spectral decomposition and localized adaptation to effectively address heterogeneous distribution shifts at test-time.
    \item We validate FreDA through extensive experiments on corrupted, natural, and medical benchmarks, demonstrating consistent improvements over state-of-the-art methods across various distribution shift scenarios.

\end{itemize}

\section{Connections to Previous Studies}

\subsection{Transfer Learning, Domain Adaptation, Test-time Training, and Test-time Adaptation}
Deep neural networks often experience performance degradation when deployed in environments different from their training settings, due to distribution shifts between source (training) and target (testing) domains. This issue has been extensively studied under the umbrella of transfer learning~\cite{pan2009survey,zhang2022transfer,TF3}, which aims to transfer knowledge across domains. A key branch of transfer learning is domain adaptation (DA), where models trained on the labeled source domain are adapted to the target domain. Depending on the data availability, previous DA can be roughly classified into: Supervised DA~\cite{SDA1} using labeled data in both source and target domains during training; Unsupervised DA~\cite{DA1,DA2,DA3,DA4}, which relies on labeled source data and unlabeled target data simultaneously; and Source-free DA~\cite{SFDA1,SFDA3,SFDA4}, that adapts to the target domain without access to source data, typically due to privacy constraints.

While traditional DA methods assume access to the entire target distribution before adaptation, real-world deployment often involves receiving data as an \emph{online stream}, where samples arrive sequentially and the full distribution is never observed. This streaming nature of the target domain breaks the assumption of global access, making conventional domain adaptation infeasible. To handle this constraint, two prominent test-time adaptation paradigms have emerged. 
Online Test-time Training (OTTT)~\cite{TTT,liu2021ttt++} introduces an auxiliary task -- often self-supervised -- during pre-training. At test time, the model optimizes this auxiliary objective to adapt to the target distribution accordingly; Online Test-time Adaptation (OTTA)~\cite{wang2021tent,niu2022towards,niu2022efficient,chen2022contrastive,wang2022continual,su2024unraveling} poses a more demanding and practical scenario, where the model must adapt on-the-fly to the test stream without any prior modification during training. This setting emphasizes the need for rapid, real-time model updates to effectively capture the continuously incoming data. 

A brief comparison of these adaptation settings is summarized in TABLE~\ref{tab:da_comparison}. 
Our work falls under the OTTA setting, where we propose a generalizable and robust adaptation framework that maintains strong performance across diverse distribution shifts and corruptions -- without relying on source data or altering the pre-training process. 

\begin{table*}[t]
\centering
\caption{Comparison of different transfer learning settings where $\checkmark$ denotes settings closer to real-world conditions.}
\label{tab:da_comparison}
\renewcommand{\arraystretch}{1.2}
\setlength\tabcolsep{7pt}
\resizebox{0.98\textwidth}{!}{
\begin{tabular}{lcccc}
\toprule
Research Fields   & No Source Data$^*$    & No Target Labels$^*$ & Online Adaptation & Model Agnostic
\\
\midrule
{Supervised Domain Adaptation} & 
$\times$ & 
$\times$ & 
$\times$ & 
--
\\
{Unsupervised Domain Adaptation} & 
$\times$ & 
$\checkmark$ & 
$\times$ & 
--
\\
{Source-free Domain Adaptation} & 
$\checkmark$ & 
$\checkmark$ & 
$\times$ & 
--
\\
{Online Test-time Training} & 
$\checkmark$ & 
$\checkmark$ & 
$\checkmark$ & 
$\times$
\\
\rowcolor{gray!10}
{Online Test-time Adaptation} & 
$\checkmark$ & 
$\checkmark$ & 
$\checkmark$ & 
$\checkmark$
\\
\bottomrule
\end{tabular}
}
\end{table*}

\subsection{Non-i.i.d. Test-time Adaptation}
While conventional TTA methods primarily focus on single domain shifts under the idealized independent and identically distributed (i.i.d.) assumption, real-world deployment often violates these conditions due to inherent data heterogeneity. This discrepancy has motivated recent efforts to extend TTA to more realistic settings, with two primary challenges:

\subsubsection{Mixed Domains} 
{Modern TTA methods designed for continual domain shifts~\cite{wang2022continual,rotta,niu2022efficient,press2024rdumb,deyo} have pushed beyond single adaptation problems. However, while they account for the dynamic target distribution shifts over time, they typically assume that each batch contains only a single domain shift. This simplifies the real-world scenario, where multiple distribution shifts may occur simultaneously, creating a mixed-domain environment that remains largely unexplored.}

Existing approaches primarily focus on stabilizing model updates through periodic parameter resets~\cite{niu2022towards,press2024rdumb} or importance weighting of target samples~\cite{niu2022efficient,press2024rdumb,deyo}, but these techniques are ill-equipped to disentangle and manage the complex interleaving of domain shifts, leading to degraded performance.
In contrast, our work explicitly addresses mixed domains via frequency-space decomposition, which facilitates domain separation and  proactive distribution alignment prior to model adaptation. While recent efforts like~\cite{niu2022towards} acknowledge mixed distribution challenges under the broader ``Dynamic Wild World'' paradigm, their unified treatment of diverse real-world factors does not systematically address data heterogeneity. Conversely, our study directly targets the core issue of mixed domains in TTA, introducing tailored strategies for disentangling and adapting to co-occurring heterogeneous distributions.

\subsubsection{Dependent Sampling} 

The second challenge arises from temporal-dependent sampling, which introduces class-level dependencies. This topic has garnered considerable research attention, with a growing body of work~\cite{rotta,gong2022note,zhao2022delta,unmixing,roid,zhou2023ods} actively addressing it through strategies such as pseudo-label-based rebalancing or extended observation windows. These approaches primarily aim to address temporal imbalances caused by skewed class distributions over time, aiming to improve target domain estimation by addressing low class diversity within a batch.
However, as these methods focus on within-domain temporal coherence, their applicability becomes limited when batches contain samples from multiple target domains. This limitation is also reflected in our experiments, where mixed-domain sampling poses notable challenges not accounted for in these approaches.

\section{Problem Definition of mixed domain shifts}
\label{sec:prob_def}
Test-time adaptation (TTA) aims to adjust a model \( q_{\mathbf{\theta}}(y|x) \), initially trained on a source dataset \( \mathcal{D}_{s} = \{(x,y) \sim p_s(x,y)\} \), to a target domain \( \mathcal{D}_{t} = \{(x,y) \sim p_t(x,y)\} \) on a data stream without accessing source data or target labels. TTA handles covariate shift by assuming \( p_s(y|x) = p_t(y|x) \) while \( p_s(x) \neq p_t(x) \). This challenge intensifies when \( \mathcal{D}_{t} \) contains multiple non-i.i.d sub-distributions \( p_{t_i}(x) \), such as
\[
p_t(x) \gets \{p_{t_1}(x), p_{t_2}(x), \dots, p_{t_N}(x)\}.
\]
Specifically, we have $\mathcal{D}_t = \mathcal{D}_{t_1} \cup \mathcal{D}_{t_2} \cup \cdots \cup \mathcal{D}_{t_N}$ where $x \in \mathcal{D}_{t_1}$ satisfying $x \sim p_{t_1}(x)$. This scenario requires the model \( q_{\mathbf{\theta}}(y|x) \) to effectively handle the heterogeneous and evolving target distribution to maintain robust performance. TTA strategies must therefore refine the model to optimize its predictive accuracy across these diverse sub-domains, ensuring consistent and reliable performance amidst significant distributional variability.

\section{Un-mixing TTA under Mixed Distribution Shifts: A Fourier Perspective}\label{sec:high_freq_effect}

\begin{figure}[t]
\centering
\includegraphics[width=0.48\textwidth,clip, trim=15cm 5.5cm 18cm 6.2cm]{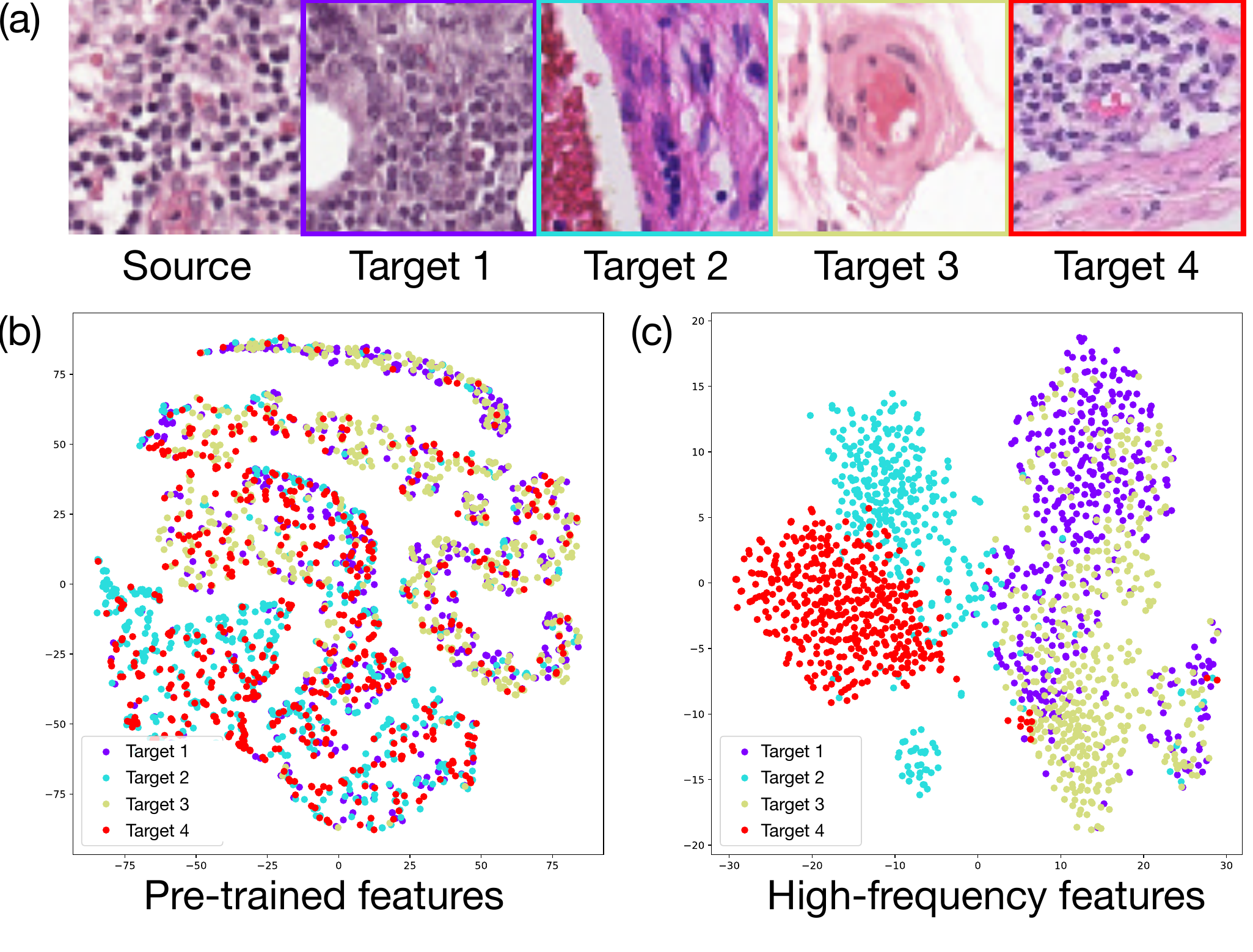}
\caption{
(a) Example histopathology patches from five distinct healthcare centers illustrate visual heterogeneity across different domains (Camelyon17~\cite{bandi2018detection}).
(b) Latent features from the pre-trained model fail to distinguish target subdomains and remain highly entangled.
(c) In contrast, high-frequency features can clearly separate different target subdomains.
}
\label{fig:high_freq_effect}
\end{figure}

While Test-Time Adaptation (TTA) methods perform well under single-type shifts, they degrade under mixed distribution shifts that are ubiquitous in real-world deployment (see Fig.~\ref{fig:tta_problem}~(b)). To address this, we advocate a paradigm of \textit{proactive distribution management}: before any parameter update, we resolve heterogeneity by decomposing the input stream in frequency space. Our key insight is that different distributions leave distinct spectral footprints, which can be utilized to disentangle the hetergenous data stream. As shown in Fig.~\ref{fig:high_freq_effect}, high-frequency components naturally separate different domain shifts, in contrast to the substantial overlap observed in the latent sample features from a pre-trained model. This proactive step transforms an ill-posed, entangled adaptation problem into a set of well-conditioned homogeneous subproblems. 

Take TBN~\cite{nado2020evaluating} as an example. In its original centralized version, TBN computes statistical information using the whole target batch and directly replaces the source values for online adaptation. While effective under single-domain shifts, this whole-batch, homogeneous strategy produces biased statistics under mixed domain shifts, leading to suboptimal adaptation (see Fig.~\ref{fig:comparedClassVisual}~(a)). In contrast, our decentralized approach first applies high-frequency clustering to partition the batch into locally homogeneous clusters. Batch normalization statistics are then computed within each cluster, enabling localized adaptation that can effectively prevent cross-domain contamination. 
Therefore, this decentralized strategy can reduce the error rate by a notable 4.27\% and yields tighter within-class cohesion along with sharper between-class boundaries compared to centralized adaptation (see Fig.~\ref{fig:comparedClassVisual}~(b)).

\section{Frequency-based Decentralized Adaptation}

As outlined in the previous section, distinguishing samples from different distribution shifts provides a promising direction for TTA. Building on this insight, we introduce Frequency-based Decentralized Adaptation (FreDA), which partitions target samples in the Fourier space into homogeneous subdomains, enabling more accurate adaptation. We further enhance this process with a augmentation strategy that diversifies each subdomain in the Fourier space to improve model robustness.

\subsection{Frequency-based Decentralized Learning}

We first introduce Frequency-based Decentralized Learning framework. It begins by extracting high-frequency information from the pixel space to partition target samples into homogeneous subsets, disclosing the latent target subdomains contained within a batch. We then perform a decentralized adaptation, where multiple local models -- sharing the same architecture -- independently adapt to their assigned subsets. At regular intervals, the parameters of these local models are aggregated to form a stronger base model, which is subsequently used to update the local models for the next adaptation phase. Detailed procedures are provided in the following subsections.

\begin{table}[t]
\label{tab:thero_evi}
\centering
\footnotesize
\setlength\tabcolsep{3pt}
\resizebox{0.46\textwidth}{!}{
\begin{tabular}{lccc}
\toprule
\textbf{Methods} & \textbf{CIFAR-10-C} & \textbf{CIFAR-100-C} & \textbf{ImageNet-C}  \\
\midrule
TBN (centralized)           & 33.8     & 45.8      & 82.5                   \\
\rowcolor{lightpink} TBN (our decentralized) & 28.5     & 43.2      & 77.6                \\
\bottomrule
\end{tabular}}
% \end{table}
\end{table}

\begin{figure}[t]
    \centering
    \begin{subfigure}[b]{0.24\textwidth}\includegraphics[width=0.96\linewidth,height=0.89\columnwidth, clip, trim=0.01cm 0cm 0cm 0.15cm]
        {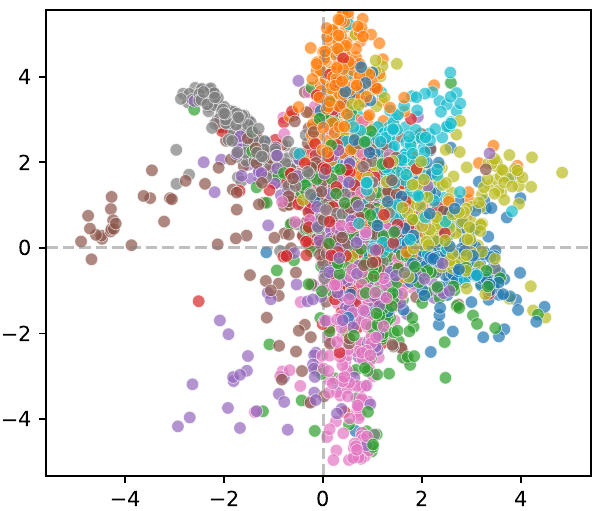}
        \caption{Centralized Adaptation}
        \label{fig:sub4}
    \end{subfigure}
    \begin{subfigure}[b]{0.24\textwidth}
\includegraphics[width=0.98\linewidth,height=0.9\columnwidth]{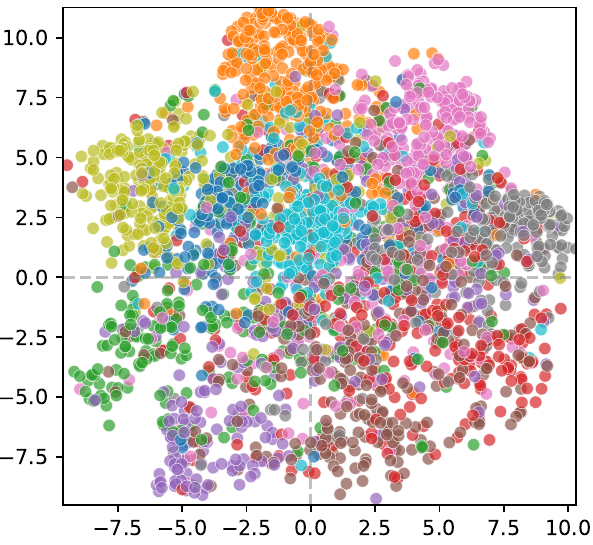}
        \caption{Decentralized Adaptation}
        \label{fig:sub3}
    \end{subfigure}
\caption{
Classification error rate (\%) and t-SNE sample feature visualization using TBN~\cite{nado2020evaluating} under mixed distribution shifts. Different colors denote different classes. (a) Original Centralized Adaptation: global Batch Normalization parameters are computed from the whole target batch. (b) Our Decentralized Adaptation: local Batch Normalization parameters are tailored for clusters separated based on high-frequency features. 
Decentralized adaptation achieves a 4.27\% error reduction and clearer class structure compared with centralized adaptation.
}
\label{fig:comparedClassVisual}
\end{figure}

\subsubsection{Frequency Feature Extraction} 
We start by extracting frequency domain features from the input images. 
Let $\mathbf{X} \in \mathbb{R}^{n \times c \times h \times w}$ denote a batch of input images, where $n$ is the batch size, $c$ is the number of channels, $h$ and $w$ are the height and width. We first apply a Fourier transform $\mathcal{F}$ to each image $\mathbf{x}_i$. This transform converts the image from the spatial domain to the frequency domain, producing a complex-valued representation \(\mathcal{F}(\mathbf{x}_i) \in \mathbb{C}^{h \times w \times c}\) that contains both real \(R(\mathbf{x}_i)\) and imaginary part \(I(\mathbf{x}_i)\). Next, we compute the amplitude spectrum \(A(\mathbf{x}_i)(u, v)\) using $A(x)(u, v) = \sqrt{R^2(x)(u, v) + I^2(x)(u, v)}$.
It reveals the intensity of the frequency content, e.g., high-frequency amplitudes highlight edges and fine details while low-frequency amplitudes emphasize the overall structure and gradual changes. Then, we filter out low-frequency elements using mask $M(u, v) = \mathbbm{1}\left( \left(u < \frac{h}{4} \lor u > \frac{3h}{4}\right) \lor \left(v < \frac{w}{4} \lor v > \frac{3w}{4}\right) \right)
\label{hf2}$
to emphasize the high-frequency components $G(x)(u, v) $ that are more likely to indicate shifts in distribution:
\begin{equation}
G(x)(u, v) = A(x)(u, v) \cdot M(u, v).
\label{hf1}
\end{equation}

\subsubsection{Frequency-Based Clustering}
We then employ a clustering algorithm (e.g., K-means) to partition the frequency features into $K$ clusters, each corresponding to a different type of distribution shift. The process is formalized as:
\begin{equation}
\min_{\mathbf{C}, \mathbf{Z}} \sum_{i=1}^{n} \left\| \mathbf{A}_{hf,i} - \mathbf{C}_{\mathbf{Z}_i} \right\|^2_2,
\label{kmeans}
\end{equation}
where $\mathbf{A}_{hf,i} = \text{vec}(G(x_i))$, $\mathbf{C} \in \mathbb{C}^{K \times d}$, $\mathbf{Z} \in \{1, \dots, K\}^n$ denotes the 1D high-frequency component of the amplitude spectrum, the centroids of the clusters and the cluster assignments for each image. $hf$ refers to high-frequency components, and $d = h \times w \times c$ is flattened dimension.

\subsubsection{Decentralized Fine-tuning}
Test-time fine-tuning is then decentralized across these clusters, allowing for specialized adaptation within each subgroup: For each cluster \( k \), we adapt a specialized model \( q_{\theta_k}(y|x) \) that is fine-tuned using only the data within that cluster:
\begin{equation}
\theta_k^* = \arg\min_{\theta_k} \mathbb{E}_{x \sim {p}_{t,k}} \left[ \mathcal{L}(q_{\theta_k}(x)) \right],
\label{update}
\end{equation}
where \( {p}_{t,k} \) represents the data distribution within cluster \( k \), and \( \mathcal{L} \) is the loss function. The predictions of each iteration are collected and sorted after the local fine-tuning.

\subsubsection{Base Model Enhancement}
To mitigate potential clustering errors for more stable adaptation model weights, we allow each local models to periodically upload their parameters for aggregation at intervals of time \textbf{$T$}:
\begin{equation}
    \mathbf{\theta}_{\text{base}} = \sum_{k=1}^{K} \left( \frac{|\mathcal{D}_k|}{\sum_{j=1}^{K} |\mathcal{D}_j|} \mathbf{\theta}_k \right),
\label{comm}
\end{equation}
\noindent where \( |\mathcal{D}_k| \) denotes sample number in cluster \( k \). This aggregation step combines the parameters from all local models in proportion to their cluster size.
The updated parameters \( \mathbf{\theta}_{\text{base}} \) are then distributed back to each subnetwork, initializing them for the next adaptation round: $\mathbf{\theta}_k \leftarrow \mathbf{\theta}_{\text{base}}$.

\subsection{Frequency-based Augmentation}

Although decentralized learning separates heterogeneous shifts into more coherent subdomains, each local model inevitably operates on a reduced set of samples. This limited within-cluster diversity may lead to unstable adaptation. To address this, we introduce a frequency-based augmentation strategy tailored for our decentralized paradigm. The method perturbs the amplitude components of target samples in the Fourier space, generating variations that are diverse yet consistent with each sample’s underlying distribution. This targeted augmentation enriches subdomain representations and strengthens model robustness across heterogeneous shifts.

\subsubsection{Sample Selection Strategy} We first select the reliable samples in each local model leveraging a criterion derived from the weighted entropy framework used in ETA~\cite{niu2022efficient} based on two primary conditions:
\begin{equation}
\text{Cri} = \mathbbm{1} \left[ \left( H(\mathbf{y}_t) < H_0 \right) \wedge \left( |\text{cos}(\mathbf{y}_t, \bar{\mathbf{y}}_{t-1})| < \epsilon \right) \right].
\label{select}
\end{equation}
The entropy
\(H(\mathbf{y}_t)\) measures the uncertainty in the current predictions.
The cosine similarity \(\text{cos}(\mathbf{y}_{t}, \bar{\mathbf{y}}_{t-1})\)  denotes the deviation between the current sample's class probabilities \(\mathbf{y}_t\) and the aggregated class probabilities \(\bar{\mathbf{y}}_{t-1}\). \(\epsilon\) is the threshold for cosine similarity, and \(H_0\) is the fixed entropy threshold.
This ensures that selected samples exhibit significant deviations from previous predictions in class distribution and lower prediction uncertainty.

\subsubsection{Sample Augmentation Mechanism} The augmentation process involves perturbing the amplitude spectrum. Let $A(x_i)$ represent the amplitude spectrum of a selected sample $x_i$. To generate a perturbed amplitude spectrum $\tilde{A}(x_i)$, we apply a random Gaussian perturbation:
\begin{equation}
\tilde{A}(x_i) = \left( 1 + \alpha \cdot \Delta  \right) \cdot A(x_i),
\label{permute}
\end{equation}
where $\Delta \sim \mathcal{N}(0, \sigma^2)$ is a perturbation matrix sampled from a Gaussian distribution, and $\alpha$ is a scaling factor.
Then, the augmented sample $\tilde{x}_i$ is reconstructed via the inverse Fourier transform $\mathcal{F}^{-1}$ to the perturbed amplitude spectrum, combined with the original phase spectrum $P(x_i)$:
\begin{equation}
\tilde{\mathbf{X}}_i = \mathcal{F}^{-1}\left(\tilde{A}(x_i), P(x_i)\right).
\label{generation}
\end{equation}

\begin{algorithm}[t]
\caption{Framework of Frequency-based Decentralized Learning and Augmentation}\label{algo:FreDA}
\footnotesize
\begin{algorithmic}[1]
\Require Step $t$, Input batch $\mathbf{X}= \{x_1, x_2,..., x_n\} \in \mathbb{R}^{n \times h \times w \times c}$, Pretrained source model $q_\theta$, Initialize Feature Repository  and Local Sample Pool $\mathcal{R}, \mathcal{S}_k \leftarrow \emptyset$, CLUSTER\_NUM $K$, KMEANS\_SIZE $N$, COMM\_INTERVAL $f$;
\Statex \textbf{\textcolor{black!40}{Step 1: Extract Frequency Features}}
\For{$i = 1$ to $n$}
    \State $\mathbf{A}_{hf,i} \leftarrow \text{vec}(\text{G}(\mathbf{x}_i))$ \Comment{\textit{Extract high-freq components}}
\EndFor
\Statex \textbf{\textcolor{black!40}{Step 2: Dynamic Clustering}}
\State $\mathcal{R} \leftarrow \mathcal{R} \cup \{\mathbf{A}_{hf,i}\}_{i=1}^n$ \Comment{\textit{Frequency Information Repository}}
\State $\mathcal{R} \leftarrow \mathcal{R}[(|\mathcal{R}| - N + 1) : ]$ 
\Comment{\textit{Keep the last $N$ entries for kmeans clustering}}
\State $(\mathbf{C}_t, \mathbf{Z}) \leftarrow \text{K-means}(\mathcal{R}, K, \mathbf{C}_{t-1})$ \Comment{\textit{Obtain Cluster Labels $\mathbf{Z} = \{Z_i\}_{i=1}^{n}$ (Eq.~\ref{kmeans})}}

\Statex \textbf{\textcolor{black!40}{Step 3: Local Model Training}}
\For {cluster $k \in \{1, \ldots, K\}$}
    \State $\mathcal{S}_k \leftarrow  \mathcal{S}_k \cup \{x_i \mid Z_i = k\}$ \Comment{\textit{Gather samples for cluster $k$}}
    \State $\mathcal{S}_k \leftarrow \mathcal{S}_k[(|\mathcal{S}_k| - n + 1) : ]$ \Comment{\textit{Keep the last $batch\_size = n$ entries}}
    \State $\mathcal{S}_k' \leftarrow \text{select\_samples}(\mathcal{S}_k)$ \Comment{\textit{Select samples (Eq.~\ref{select})}}
    \For{each $x_i \in \mathcal{S}_k'$}
        \State $\tilde{x}_i \leftarrow \text{augment}(x_i)$ \Comment{\textit{Augment data (Eq.~\ref{generation})}}
        \State $\text{Train}(q_{\theta_k},x_i, \tilde{x}_i)$ \Comment{\textit{Train local model (Eq.~\ref{update})}}
    \EndFor
\EndFor

\Statex \textbf{\textcolor{black!40}{Step 4: Compile Predictions}}
\State $\mathbf{Y} \leftarrow \text{collect\_sort}(\{q_{\theta_k}(\mathbf{X})\})$ \Comment{\textit{Collect and sort predictions}}

\Statex \textbf{\textcolor{black!40}{Step 5: Base Model Enhancement}}
\State If t \% $f$ == 0 : \Comment{\textit{Model Communication with interval $f$ (Eq.\ref{comm})}}

\State \quad $\mathbf{\theta}_{\text{global}} \leftarrow \sum_{k=1}^K w_k \theta_k$ 
\State     \quad$\mathbf{\theta}_k \leftarrow \mathbf{\theta}_{\text{global}}$ 

\end{algorithmic}
\end{algorithm}

\subsubsection{Loss Function} The training objective combines the entropy loss of the selected samples with a consistency loss from the augmented samples. The total loss is defined as:
\begin{equation}
\label{lossall}
\mathcal{L}_{\text{total}} = \frac{1}{n} \sum_{i=1}^{n} H(\mathbf{y}_i) + \lambda \cdot \frac{1}{n} \sum_{i=1}^{n} \mathcal{L}_{\text{con}}\left(\hat{\mathbf{y}}_i, \tilde{\mathbf{y}}_i\right),
\end{equation}
where the entropy loss $H(\mathbf{y}_i)$ for the original sample $x_i$ is given by
$H(\mathbf{y}_i) = - \sum_{j=1}^{C} \mathbf{y}_{i,j} \log \mathbf{y}_{i,j}$
with $\mathbf{y}_i$ being the predicted probability over the $C$ classes, and the consistency loss 
% $\mathcal{L}_{\text{con}}\left(\tilde{\mathbf{y}}_i, \hat{\mathbf{y}}_i\right)$ 
$\mathcal{L}_{\text{con}}\left(\hat{\mathbf{y}}_i, \tilde{\mathbf{y}}_i\right) = - \sum_{j=1}^{C} \hat{\mathbf{y}}_{i,j} \log \tilde{\mathbf{y}}_{i,j}$
is defined as the cross-entropy between the prediction $\tilde{\mathbf{y}}_i$ of the augmented sample $\tilde{x}_i$ and the pseudo-label $\hat{\mathbf{y}}_i$ from the original sample.

\subsection{Overall Framework}
{We provide the overall pipeline algorithm of FreDA in Algorithm~\ref{algo:FreDA}. During implementation, we leverage a memory bank strategy~\cite{rotta,chen2022contrastive,zhang2023adanpc,karmanov2024TDA} that is updated in real time. This design serves two purposes: \textbf{1)} to ensure accurate clustering -- since an overly small batch could impede the effective separation of data -- and \textbf{2)} to maintain the number of samples processed by the local model consistent with the original batch size, thereby preventing performance degradation due to a drastic reduction in batch size (e.g., reducing by a factor of the cluster number).}

\section{Theoretical Insights}

To facilitate theoretical understanding of our method, we adopt the expansion-based framework~\cite{wei2020theoretical}, which offers a principled analysis of pseudo-labeling under local perturbations and neighborhood consistency, and connects these properties directly to upper bounds on model error rates. 
We first revisit the key definitions and theorems, then demonstrate how our method effectively tightens this generalization bound under mixed domain shifts from the viewpoints of \emph{Frequency-coherent Partition} and \emph{Augmentation-induced Expansion}.

\subsection{Expansion Theory}

\begin{definition}[$(a,c)$-expansion]
A class-conditional distribution $P_i$ satisfies $(a,c)$-expansion if $\forall S \subseteq \mathcal{X}$ with $P_i(S) \leq a$:
\begin{equation}
\mathcal{P}_i(\mathcal{N}(S)) \geq \min(c\mathcal{P}_i(S), 1)
\nonumber
\end{equation}
where $\mathcal{N}(S)$ refers to the neighborhood of $S$ under data augmentations.
\end{definition}

\begin{definition}[Separation]
$P$ is $(\mu,r)$-separated if:
\begin{equation}
\mathbb{E}_{x\sim\mathcal{P}}\left[\max_{x' \in B_r(x)} \mathbf{1}(G(x) \neq G(x'))\right] \leq \mu
\nonumber
\end{equation}
where $B_r(x)$ is an $\ell_2$-ball of radius $r$.
\end{definition}

\noindent The core theorem from \cite{wei2020theoretical} states:

\begin{theorem}[Pseudo-label Denoising]
\label{thm:wei}
Under $(a,c)$-expansion and $(\mu,r)$-separation, any classifier $G$ of:
\begin{equation}
\min_G \frac{c+1}{c-1}\mathcal{L}_{pl}(G) + \frac{2c}{c-1}\mathcal{R}_B(G)
\nonumber
\end{equation}
achieves error:
\begin{equation}
\mathrm{Err}(G) \leq  \frac{2}{c-1}\mathrm{Err}(G_{pl}) +\frac{2c}{c-1}\mu 
\nonumber
\end{equation}
% where 
$\mathcal{L}_{pl}$ is pseudo-label loss and $\mathcal{R}_B$ is consistency regularizer.
\end{theorem}

\subsection{Analysis of Our Method}
Our method reduces both terms in Theorem~\ref{thm:wei} to achieve a tighter bound from two perspectives:

\subsubsection{Frequency-based Partition}
Let $\{P_{t_k}\}_{k=1}^{K}$ be the $K$ sub-domains derived from frequency-based partitioning. For the related measurable sets $\{S_k\}_{k=1}^{K}$ ($S_k \subseteq X_k$),
% (the support of $P_{t_k}$),
we have:

\begin{equation}
\mathrm{Err}(G_{pl}^k) \leq \mathrm{Err}(G_{pl}) - \Delta_k \quad
\nonumber
\end{equation}

\noindent where $\Delta_k = \mathcal{P}_{t_k}\left( E_c\right)$ and $E_c = \{ x \in X_k \mid G_{pl}(x) \neq y(x) \}$. This contributes to the reduction of the first term in Theorem~\ref{thm:wei}, which is further supported by experimental results in Fig.~\ref{fig:comparedClassVisual}.

\subsubsection{Augmentation-Induced Expansion}
Our frequency augmentation expands neighborhoods:
\begin{equation}
\widehat{\mathcal{N}}(S) = \mathcal{N}_{\text{base}}(S) \cup \left\{x': \inf_{x \in S} |A(\mathcal{F}(x')) - A(\mathcal{F}(x))|_2 \leq \epsilon \right\}
\nonumber
\end{equation}

\noindent where $\mathcal{N}_{\text{base}}(S)$ is the base augmentation neighborhood. It is worth noting that, unlike most methods, we do not apply base augmentation.  The term  $\mathcal{N}_{\text{base}}(S)$ is retained in the formulation for theoretical completeness. Our choice of this representation is intended to highlight the expansion gain brought by incorporating frequency augmentation. The expansion gain becomes:
\begin{equation}
\widehat{c} = c_{\text{base}} + \gamma,\ 
\gamma = \inf_{S:P(S)\leq a} \frac{P(\widehat{\mathcal{N}}(S)\setminus\mathcal{N}_{\text{base}}(S))}{P(S)}
\nonumber
\end{equation}

\noindent Substituting into Theorem~\ref{thm:wei}, the error becomes:
\begin{equation}
\begin{split}
\mathrm{Err}(G) &\leq  \underbrace{\frac{2}{(c_{\text{base}}+\gamma)-1}\left(\mathrm{Err}(G_{pl}) - \sum_{k=1}^{K} \Delta_k\right)}_{\text{Reduced by both terms}} \\ &+\underbrace{\frac{2(c_{\text{base}}+\gamma)}{(c_{\text{base}}+\gamma)-1}\mu}_{\text{Reduced by } O(\gamma/c^2)} 
\nonumber
\end{split}
\end{equation}

\noindent \textbf{Remark.} By leveraging frequency-based decentralized learning and augmentation, our method addresses the key components in the expansion-theoretic error, yielding a provably tighter generalization bound that supports more challenging online adaptation scenarios under diverse distribution shifts.

\noindent\textbf{Notation}
\begin{itemize}
\item $c$: Expansion factor (Def. 1). 

\emph{Class-wise connectivity metric: Larger $c$ implies stronger neighborhood propagation of local consistency to global predictions.}

\item $\mu$: Inter-class separation probability (Def. 2).

\emph{Robustness measure: Probability of different classes having overlapping neighborhoods under perturbations, lower $\mu$ indicates clearer decision boundaries.}

\item $\mathcal{R}_B$: Consistency regularizer (Thm. 1). 

\emph{Stability term: Penalizes prediction inconsistency between original inputs and their augmented variants.}

\item $\Delta_k$: Error reduction in $k$-th sub-domain. 

\emph{Specialization gain: Reduced pseudo-label noise in sub-domain $k$ due to frequency-based partitioning.}

\item $\gamma$: Expansion gain from frequency augmentation. 

\emph{Neighborhood enhancement: Additional expansion capability measured as the relative increase of augmented neighborhoods.}
\end{itemize}

\begin{table*}[t]
\centering
\caption{Classification error rate ($\downarrow$) on CIFAR-10-C, CIFAR-100-C, ImageNet-C  under \textbf{Mixed Distribution Shifts}.}
\label{tab:performance_evaluation}
\renewcommand{\arraystretch}{0.9}
\footnotesize
\resizebox{\textwidth}{!}{
\begin{tabular}{lcccccccccccccccc}
\toprule
 \textbf{Methods} & \makecell{Gauss.} & \makecell{Shot} & \makecell{Impul.} & \makecell{Defoc.} & \makecell{Glass} & \makecell{Motion} & \makecell{Zoom} & \makecell{Snow} & \makecell{Frost} & \makecell{Fog} & \makecell{Brig.} & \makecell{Contr.} & \makecell{Elast.} & \makecell{Pixel} & \makecell{JPEG} & \cellcolor{lightblue}\textbf{Avg.} \\ 
\midrule
\textbf{CIFAR-10-C} & 72.3 & 65.7 & 72.9 & 46.9 & 54.3 & 34.8 & 42.0 & 25.1 & 41.3 & 26.0 & \underline{9.3} & 46.7 & \textbf{26.6} & 58.4 & 30.3 & \cellcolor{lightblue}43.5 \\
TBN & 45.5 & 42.8 & 59.7 & 34.2 & 44.3 & 29.8 & 32.0 & 19.8 & 21.1 & 21.5 & \underline{9.3} & 27.9 & 33.1 & 55.5 & 30.8 & \cellcolor{lightblue}33.8 \\
TENT & 73.5 & 70.1 & 81.4 & 31.6 & 60.3 & 29.6 & 28.5 & 30.8 & 35.3 & 25.7 & 13.6 & 44.2 & 32.6 & 70.2 & 34.9 & \cellcolor{lightblue}44.1 \\
ETA & 36.2 & 33.3 & 52.3 & 22.9 & 38.9 & 22.4 & 20.5 & 19.5 & 19.7 & 20.4 & 11.3 & 35.4 & \textbf{26.6} & 38.8 & \underline{25.1} & \cellcolor{lightblue}28.2 \\
AdaContrast & 36.7 & 34.3 & 48.8 & \textbf{18.2} & 39.1 & 21.1 & \underline{17.7} & \underline{18.6} & \underline{18.3} & \underline{16.8} & \textbf{9.0} & \underline{17.4} & 27.7 & 44.8 & \textbf{24.9} & \cellcolor{lightblue}\underline{26.2} \\
CoTTA  & 38.7 & 36.0 & 56.1 & 36.0 & \textbf{36.8} & 32.3 & 31.0 & 19.9 & \textbf{17.6} & 27.2 & 11.7 & 52.6 & 30.5 & \underline{35.8} & 25.7 & \cellcolor{lightblue}32.5 \\
SAR  & 45.5 & 42.7 & 59.6 & 34.1 & 44.3 & 29.7 & 31.9 & 19.8 & 21.1 & 21.5 & \underline{9.3} & 27.8 & 33.0 & 55.4 & 30.8 & \cellcolor{lightblue}33.8 \\
RoTTA  & 60.0 & 55.5 & 70.0 & 23.8 & 44.1 & \underline{20.7} & 21.3 & 20.2 & 22.7 & \textbf{16.0} & 9.4 & 22.7 & 27.0 & 58.6 & 29.2 & \cellcolor{lightblue}33.4 \\
RDumb & \underline{34.9} & \underline{32.3} & 49.4 & 23.3 & \underline{38.2} & 23.3 & 20.7 & 19.9 & 19.3 & 20.7 & 11.2 & 29.3 & \underline{26.7} & 41.5 & 25.2 & \cellcolor{lightblue}27.7 \\
DeYO  & 45.8 & 42.3 & 65.7 & 21.3 & 41.8 & 25.1 & 19.5 & 21.1 & 19.6 & 19.2 & 12.3 & 21.8 & 28.5 & 39.3 & 28.0 & \cellcolor{lightblue}30.1 \\
UnMix-TNS & 50.0 & 44.4 & \underline{44.3} & 34.4 & 48.2 & 32.7 & 30.0 & 35.5 & 35.9 & 47.5 & 28.1 & 38.7 & 43.9 & 40.0 & 43.3 & \cellcolor{lightblue}39.8 \\
\rowcolor{lightpink} \textbf{FreDA (ours)} & \textbf{23.1} & \textbf{22.2} & \textbf{32.2} & \underline{18.7} & 41.6 & \textbf{18.8} & \textbf{16.8} & \textbf{17.9} & 19.9 & 16.9 & 9.8 & \textbf{13.2} & 29.1 & \textbf{35.4} & 28.6 & \textbf{22.9} \\
\midrule
\textbf{CIFAR-100-C} & 73.0 & 68.0 & 39.4 & 29.3 & 54.1 & 30.8 & \underline{28.8} & 39.5 & 45.8 & 50.3 & 29.5 & 55.1 & 37.2 & 74.7 & 41.2 & \cellcolor{lightblue}46.4 \\
TBN & 62.7 & 60.7 & 43.1 & 35.5 & 50.3 & 35.7 & 34.4 & 39.9 & 51.5 & \textbf{27.5} & 45.5 & 42.3 & 72.8 & 46.4 & 45.8 & \cellcolor{lightblue}45.8 \\
TENT & 95.6 & 95.2 & 89.2 & 72.8 & 82.9 & 74.4 & 72.3 & 78.0 & 79.7 & 84.7 & 71.0 & 88.5 & 77.8 & 96.8 & 78.7 & \cellcolor{lightblue}82.5 \\
ETA & 42.6 & 40.3 & \textbf{34.1} & 30.3 & \underline{42.4} & 32.0 & 29.4 & \underline{35.6} & \underline{35.8} & 44.1 & 30.2 & 41.8 & \textbf{36.9} & 38.9 & 40.9 & \cellcolor{lightblue}37.0 \\
AdaContrast & 54.5 & 51.5 & 37.6 & 30.7 & 45.4 & 32.1 & 30.3 & 36.9 & 36.5 & 45.3 & \underline{28.0} & 42.7 & 38.2 & 75.4 & 41.7 & \cellcolor{lightblue}41.8 \\
CoTTA & 54.4 & 52.7 & 49.8 & 36.0 & 45.8 & 36.7 & 33.9 & 38.9 & \underline{35.8} & 52.0 & 30.4 & 60.9 & 40.2 & \underline{38.0} & 41.1 & \cellcolor{lightblue}43.1 \\
SAR & 75.8 & 72.7 & 41.1 & \textbf{29.2} & 45.2 & \underline{31.1} & {28.9} & 36.7 & 37.7 & 43.9 & 29.3 & 41.8 & \underline{37.1} & 89.2 & 42.4 & \cellcolor{lightblue}45.5 \\
RoTTA & 65.0 & 62.3 & 39.3 & 33.4 & 50.0 & 34.2 & 32.6 & 36.6 & 36.5 & 45.0 & \textbf{26.4} & \underline{41.6} & 40.6 & 89.5 & 48.5 & \cellcolor{lightblue}45.4 \\
RDumb  & \underline{42.3} & \underline{40.0} & \textbf{34.1} & 30.5 & \underline{42.4} & 31.9 & 29.5 & 35.7 & 35.9 & 43.6 & 30.4 & 41.9 & {36.9} & 38.1 & \underline{40.5} & \cellcolor{lightblue}\underline{36.9} \\
DeYO & 57.2 & 53.4 & 38.8 & 34.7 & 47.3 & 37.3 & 34.1 & 40.8 & 40.5 & 50.6 & 33.3 & 45.8 & 41.5 & 94.5 & 45.7 & \cellcolor{lightblue}46.4 \\
UnMix-TNS & 65.8 & 64.1 & 46.4 & 37.5 & 51.7 & 36.0 & 36.4 & 38.5 & 39.4 & 51.1 & 29.3 & 42.8 & 43.2 & 67.8 & 49.4 & \cellcolor{lightblue}46.6 \\
\rowcolor{lightpink} \textbf{FreDA (ours)} & \textbf{34.8} & \textbf{34.7} & \underline{36.6} & \underline{29.4} & \textbf{41.2} & \textbf{29.9} & \textbf{28.4} & \textbf{33.8} & \textbf{33.7} & \underline{41.1} & 29.8 & \textbf{34.9} & \textbf{36.9} & \textbf{37.1} & \textbf{38.7} & \textbf{34.7} \\
\midrule
\textbf{ImageNet-C} & 97.8 & 97.1 & 98.2 & 81.7 & 89.8 & 85.2 & 77.9 & 83.5 & 77.1 & 75.9 & \underline{41.3} & 94.5 & 82.5 & 79.3 & 68.6 & \cellcolor{lightblue}82.0 \\
TBN & 92.8 & 91.1 & 92.5 & 87.8 & 90.2 & 87.2 & 82.2 & 82.2 & 82.0 & 79.8 & 48.0 & 92.5 & 83.5 & 75.6 & 70.4 &\cellcolor{lightblue}82.5 \\
TENT & 99.2 & 98.7 & 99.0 & 90.5 & 95.1 & 90.5 & 84.6 & 86.6 & 84.0 & 86.5 & 46.7 & 98.1 & 86.1 & 77.7 & 72.9 & \cellcolor{lightblue}86.4 \\
ETA & 90.7 & 89.2 & 90.5 & \underline{77.0} & \textbf{80.6} & \underline{74.0} & \underline{68.9} & \underline{72.4} & \underline{70.3} & \underline{64.6} & 43.9 & 93.4 & \textbf{69.2} & \textbf{52.3} & \textbf{55.9} & \cellcolor{lightblue}\underline{72.9} \\
AdaContrast  & 96.2 & 95.5 & 96.2 & 93.2 & 96.4 & 96.3 & 90.5 & 92.7 & 91.9 & 92.4 & 50.8 & 97.0 & 96.6 & 89.7 & 87.1 & \cellcolor{lightblue}90.8 \\
CoTTA  & 89.1 & \underline{86.6} & \underline{88.5} & 80.9 & 87.2 & 81.1 & 75.8 & 73.3 & 75.2 & 70.5 & 41.6 & \underline{85.0} & 78.1 & 65.6 & 61.6 & \cellcolor{lightblue}76.0 \\
SAR  & 98.4 & 97.3 & 98.0 & 84.0 & 87.3 & 82.6 & 77.2 & 77.5 & 76.1 & 72.5 & 43.1 & 96.0 & 78.3 & 61.8 & 60.4 & \cellcolor{lightblue}79.4 \\
RoTTA  & 89.4 & 88.6 & 89.3 & 83.4 & 89.1 & 86.2 & 80.0 & 78.9 & 76.9 & 74.2 & \textbf{37.4} & 89.6 & 79.5 & 69.0 & 59.6 & \cellcolor{lightblue}78.1 \\
RDumb  & \underline{89.0} & 87.6 & 88.6 & {78.1} & 82.3 & 75.2 & 70.1 & 73.0 & 71.0 & 65.1 & 43.9 & 92.6 & \underline{70.7} & \underline{53.7} & \underline{56.3} & \cellcolor{lightblue}73.1 \\
DeYO  & 99.5 & 99.2 & 99.5 & 89.5 & 95.0 & 83.9 & 78.8 & 75.0 & 87.8 & 79.2 & 47.3 & 99.2 & 92.4 & 59.0 & 60.4 & \cellcolor{lightblue}83.0 \\
UnMix-TNS & 91.7 & 92.8 & 91.7 & 92.3 & 93.4 & 91.5 & 84.8 & 86.3 & 84.1 & 85.0 & 62.0 & 96.5 & 88.6 & 81.7 & 77.3 & \cellcolor{lightblue}86.7 \\
\rowcolor{lightpink} \textbf{FreDA (ours)} & \textbf{72.4} & \textbf{74.0} & \textbf{71.4} & \textbf{76.5} & \underline{82.3} & \textbf{72.1} & \textbf{64.1} & \textbf{64.4} & \textbf{64.8} & \textbf{59.1} & 43.7 & \textbf{79.7} & {71.0} & 54.2 & 58.6 & \textbf{67.2} \\
\bottomrule
\end{tabular}
}
\caption*{
\small{
Note: WRN-28, ResNeXt-29, and ResNet-50 backbones are used for CIFAR-10-C, CIFAR-100-C, and ImageNet-C, respectively.
}
}
\end{table*}

\section{Experiments}

\subsection{Datasets and Experimental Setup} 
\subsubsection{Datasets} 
To provide a comprehensive evaluation of TTA deployment, we test models under three different scenarios (see Fig.~\ref{fig:high_freq_effect} and Appendix~\ref{apdix:data_visual} for visualizations):
\begin{itemize}
    \item Common Image Corruptions: We evaluate models on CIFAR-10-C, CIFAR-100-C, and ImageNet-C~\cite{imagenetc} with 10, 100 and 1000 classes, respectively. These benchmarks are designed to assess the model robustness against various corruptions. Each dataset consists of 15 distinct corruptions across five severity levels, resulting in 150,000 at each severity for CIFAR-10-C/100-C, and 750,000 for ImageNet-C.
    \item Natural Domain Shifts: We extend evaluation to DomainNet126~\cite{domainnet126}, which presents natural shifts across four domains (Real, Clipart, Painting, Sketch) encompassing 126 classes as a subset of the larger DomainNet dataset.
    \item Medical Application: Models are further evaluated on  Camelyon17~\cite{bandi2018detection}, comprising over 450,000 histopathological  patches from lymph node sections for binary classification of normal and tumor tissue, with data originating from five distinct healthcare centers.
\end{itemize}
For corruption datasets, the model is pretrained on the clean dataset and the 15 corruptions are randomly mixed as the target distribution. We leverage the highest severity = 5 in all the experiments. In DomainNet126 and Camelyon17, one subdomain is selected as the source, and the others serve as mixed target distributions. All reported results are averaged over runs with fixed seeds (0, 1, and 2).

\subsubsection{Baselines}
We compare FreDA with 10 models, including TBN~\cite{nado2020evaluating},
TENT~\cite{wang2021tent},
CoTTA~\cite{wang2022continual},
ETA~\cite{niu2022efficient},
SAR~\cite{niu2022towards},
AdaContrast~\cite{chen2022contrastive},
RoTTA~\cite{rotta},
RDumb~\cite{press2024rdumb},
DeYO~\cite{deyo}, and
UnMix-TNS~\cite{unmixing}. 
TBN~\cite{nado2020evaluating} re-estimates batch normalization statistics from test data.
TENT~\cite{wang2021tent} minimizes prediction entropy to optimize batch normalization.
CoTTA~\cite{wang2022continual} addresses long-term test-time adaptation in changing environments. 
ETA~\cite{niu2022efficient} and SAR~\cite{niu2022towards} exclude unreliable and redundant samples during optimization. 
AdaContrast~\cite{chen2022contrastive} utilizes contrastive learning to refine pseudo-labels and improve feature learning.  
RoTTA~\cite{rotta} presents a robust batch normalization scheme with a memory bank for category-balanced estimation.
RDumb~\cite{press2024rdumb} leverages weighted entropy and periodically resets the model to its pretrained state to prevent collapse.
DeYO~\cite{deyo} quantifies the impact of object-destructive transformations for sample selection and weighting.
UnMix-TNS~\cite{unmixing} introduces a test-time normalization layer for non-i.i.d. environments by decomposing BN statistics.
For fair comparisons, 
we conduct experiments using the open source online TTA repository~\cite{dobler2023robust}\footnote{\url{https://github.com/mariodoebler/test-time-adaptation}}, which provides codes and configurations of state-of-the-art TTA methods.

\subsubsection{Hyperparameter Configuration} 
The adaptation batch size is set to  200, 64, 128 and 32 for  CIFAR-10/100-C, ImageNet-C, DomainNet126 and  Camelyon17 following the previous methods. The SGD optimizer is used with learning rates adjusted to  0.01, 0.0001,  0.001 and  0.00005,  respectively. 
The learning rate is proportionally decreased in the experiment studying the effect of batch size. 
The Kmeans Size is 512, Clutser Number is 4, Communication Interval is 10 across all the tasks. The perturbation magnitude $\alpha$ is fixed to 0.1 and the coefficient $\lambda$ in loss function is fixed to 0.5. 
Two threshold in Eq.~\ref{select} is set to the same value for corruption datasets and DomainNet126 following ETA~\cite{niu2022efficient}. While for Camelyon17, the class diversity related threshold is adjusted to 0.9 empirically. 

\begin{table*}[t]
\setlength{\tabcolsep}{8pt}
\centering
\caption{Classification error rate ($\downarrow$) on DomainNet126 and Camelyon17 under  \textbf{Mixed Distribution Shifts}.} 
\renewcommand{\arraystretch}{0.92}
\begin{tabularx}{\textwidth}{XX}
\toprule
\begin{minipage}[t]{0.25\textwidth}
\footnotesize
\centering
% \textbf{Domainnet126}
\begin{tabular}{lccccc}
\multicolumn{6}{c}{\textbf{DomainNet126}} \\ \midrule
\textbf{Methods} & Real & Painting & Clipart & Sketch & \cellcolor{lightblue}\textbf{Avg.} \\
\midrule
Source & 45.2 & 41.6 & 49.5 & 45.3 & \cellcolor{lightblue}45.4 \\
TBN & 45.5 & 39.9 & 45.9 & 37.5 & \cellcolor{lightblue}42.2 \\
{TENT } & 42.2 & 37.8 & 44.7 & 37.5 & \cellcolor{lightblue}40.6 \\
{ETA } & 41.1 & 37.3 & \underline{43.4} & \underline{36.4} & \cellcolor{lightblue}\underline{39.5} \\
{SAR } & 43.2 & 38.5 & 44.8 & 37.0 & \cellcolor{lightblue}40.9 \\
{DeYO } & \underline{40.9} & \underline{36.4} & 43.6 & 36.9 & \cellcolor{lightblue}39.4 \\
\cellcolor{lightpink}\textbf{FreDA (ours)} & \cellcolor{lightpink}\textbf{40.2} & \cellcolor{lightpink}\textbf{36.1} & \cellcolor{lightpink}\textbf{40.0} & \cellcolor{lightpink}\textbf{33.6} & \cellcolor{lightpink}\textbf{37.5} \\
\end{tabular}
\end{minipage}
&
\begin{minipage}[t]{0.5\textwidth}
\footnotesize
\centering
\setlength{\tabcolsep}{11pt}
\begin{tabular}{cccccc}
\multicolumn{6}{c}{\textbf{Camelyon17}} \\
\midrule
 C1 & C2 & C3 & C4 & C5 & \cellcolor{lightblue}\textbf{Avg.} \\
\midrule
     \underline{21.6} & 43.6 & 52.5 & 47.4 & 47.6 & \cellcolor{lightblue}42.5 \\
     26.5 & \underline{38.5} & \underline{31.7} & \textbf{39.4} & \underline{32.8} & \cellcolor{lightblue}\underline{33.8} \\
      44.7 & 50.5 & 49.9 & 49.1 & 48.6 & \cellcolor{lightblue}48.6 \\
       47.4 & 52.5 & 47.9 & 49.9 & 39.2 &\cellcolor{lightblue}47.4 \\
     26.5 & \underline{38.5} & \underline{31.7} & \textbf{39.4} & \underline{32.8} &\cellcolor{lightblue}\underline{33.8} \\
     50.4 & 50.3 & 48.8 & 51.7 & 50.5 &\cellcolor{lightblue}50.4 \\
 \cellcolor{lightpink}\textbf{18.6} & \cellcolor{lightpink}\textbf{24.7} & \cellcolor{lightpink}\textbf{24.8} & \cellcolor{lightpink}\underline{40.5} & \cellcolor{lightpink}\textbf{30.8} & \cellcolor{lightpink}\textbf{27.9} \\
\end{tabular}
\end{minipage} \\
\bottomrule
\end{tabularx}
\label{tab:performance_evaluation2}
\end{table*}

\begin{figure}[t]
\centering
\includegraphics[width=0.95\linewidth]{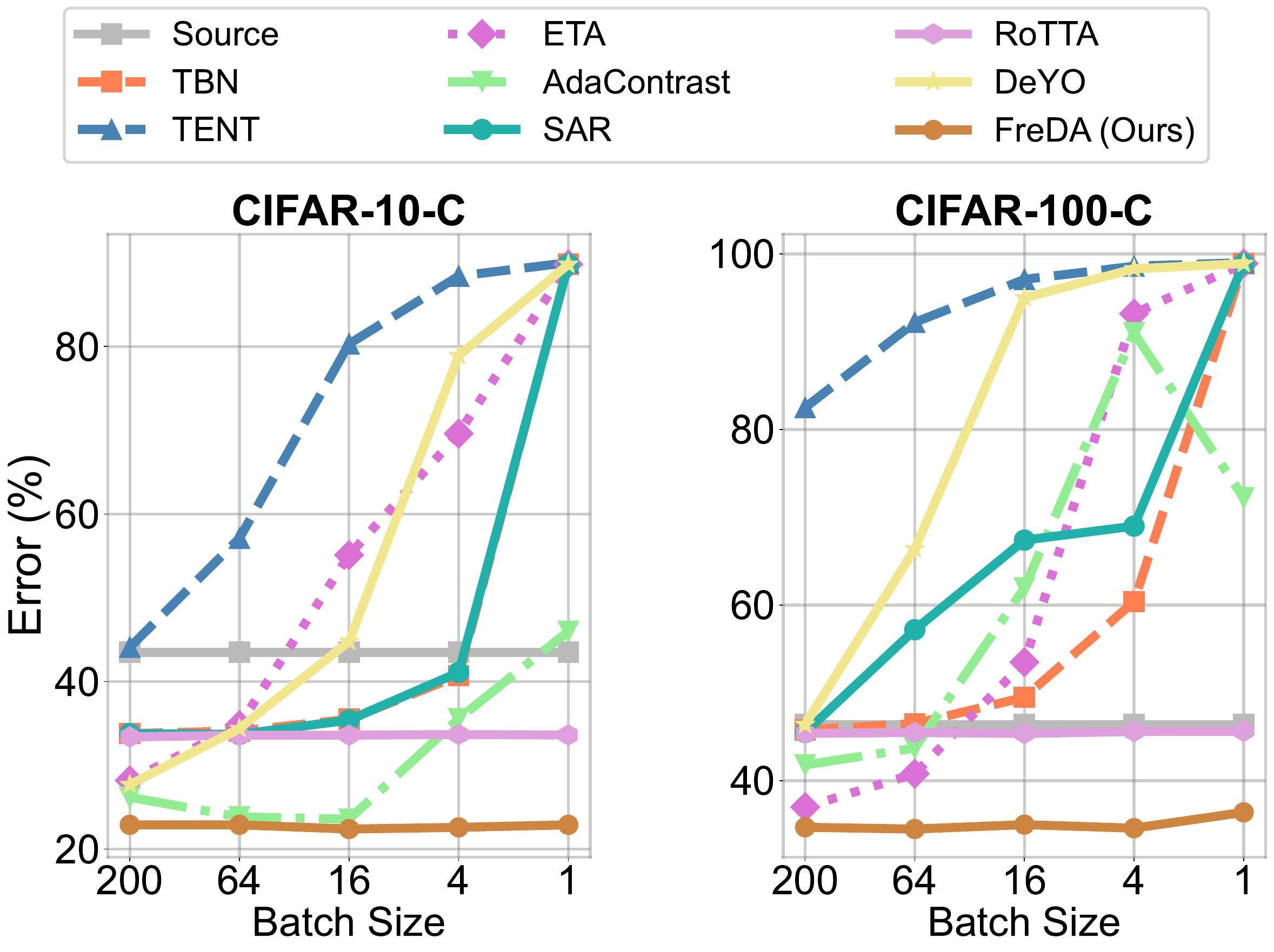}
\caption{
Classification error rate (↓) on CIFAR-10-C/100-C with various batch size under \textbf{Mixed Distribution Shifts}. 
}\label{fig:batch_size}
\end{figure}

\subsubsection{Implementation Details}
We utilize models from RobustBench~\cite{croce2021robustbench}, including WildResNet-28~\cite{wideresnet} for CIFAR-10-C and ResNeXt-29~\cite{resnext} for CIFAR-100-C, both pretrained by~\cite{hendrycks2019augmix}. For ImageNet-C, the pretrained ResNet-50~\cite{he2016deep}  
is obtained from \texttt{torchvision}. For DomainNet126, pretrained ResNet-50 is sourced from AdaContrast~\cite{chen2022contrastive}, while for Camelyon17, we train a DenseNet-121~\cite{huang2017densely} from scratch to 100 epochs with other training specifications outlined in the Wilds benchmark~\cite{koh2021wilds}.
Our implementation can be found at \url{https://github.com/kiwi12138/FreDA}.

\subsection{Main Results}
\subsubsection{FreDA improves across diverse distribution shifts}
Our method consistently attains the lowest error rates across all evaluated datasets (see TABLE~\ref{tab:performance_evaluation} and~\ref{tab:performance_evaluation2}). 
Notably, on the Camelyon17 dataset, FreDA reduced the error rate to 27.9\%, outperforming the next best method by 5.9\%. This significant improvement is particularly notable where other approaches falter if compared with no training (TBN), meaning they struggle to adapt to the complex medical imaging data. By effectively handling high variability and intricate patterns in the data, FreDA maintains superior accuracy and adaptability.

\subsubsection{FreDA remains stable under various batch size} To simulate  deployment with constrained batch sizes, we evaluate models under both varying batch sizes and mixed distributions. In Fig.~\ref{fig:batch_size}, we present the results on CIFAR-10-C and CIFAR-100-C using batch sizes ranging from  200 down to 1. Unlike other methods that significantly degrade as batch size decreases -- for example the error rate of DeYO increases from 27.7\% to 89.8\% when batch size drops from 200 to 1 on CIFAR-10-C -- FreDA consistently maintains strong performance. This stability demonstrates FreDA's robustness, making it highly suitable for real-world applications where large batches is not always feasible.

\subsubsection{FreDA enhances adaptation via frequency-based designs} 
\begin{figure}[t]
\centering
\includegraphics[width=0.99\linewidth]{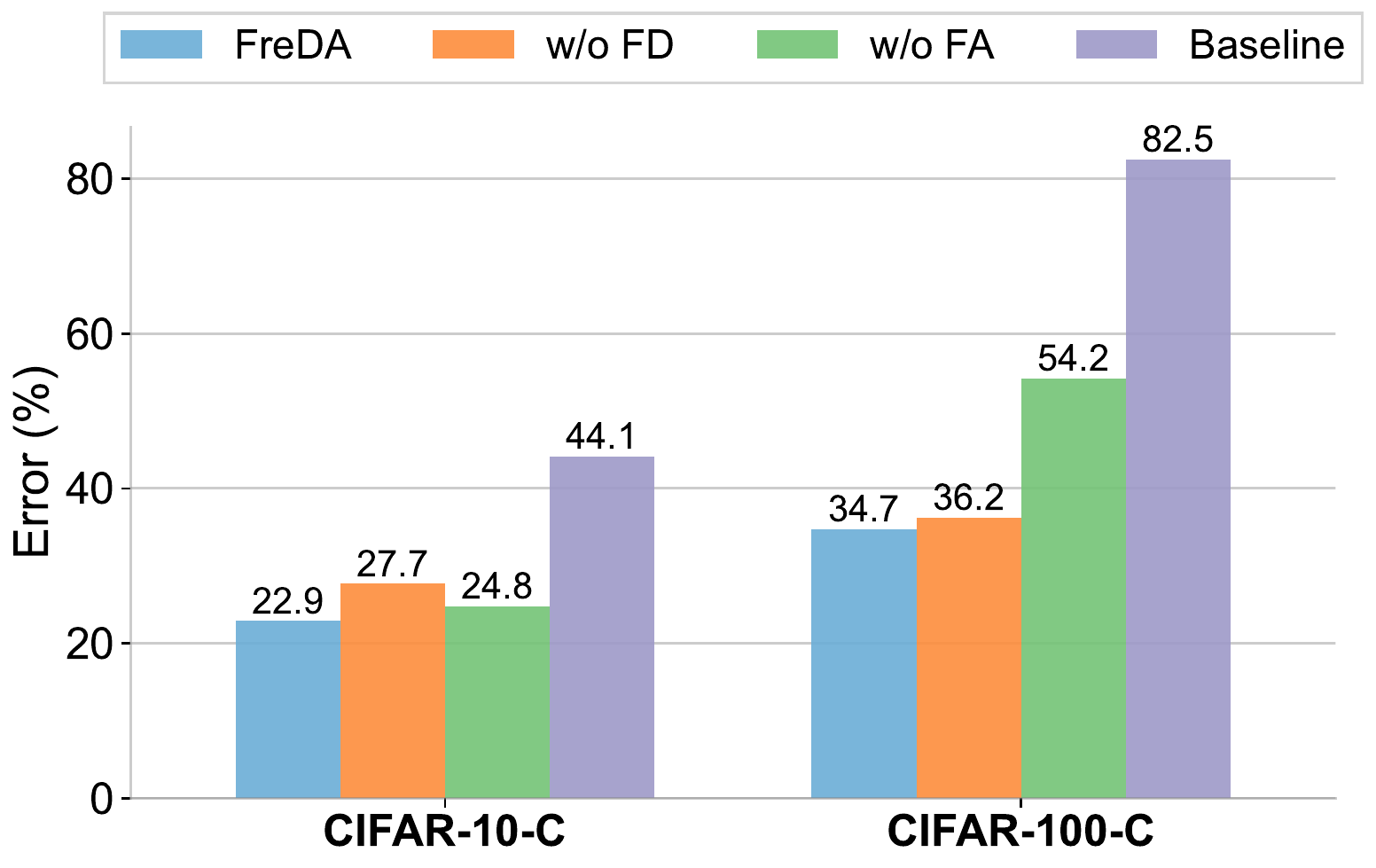}
\caption{
Classification error rate (↓) on CIFAR-10-C/100-C with various batch size under \textbf{Mixed Distribution Shifts}. 
}\label{fig:ablation}
\end{figure}

This section validates our designs by ablating two key modules -- Frequency-based Decentralized learning (FD) and Frequency-based Augmentation (FA) -- from the full model. It is worth mentioning that our FreDA is built upon TENT (our baseline) by integrating both FD and FA. From Fig.~\ref{fig:ablation}, we observe the following: 
\textbf{a)} Removing FD leads to a clear performance drop, underscoring the importance of our proposed decentralized adaptation paradigm in handling mixed distribution shifts. 
\textbf{b)} Removing FA results in a surprisingly large error rate increase on CIFAR-100-C. In more challenging classification scenarios (e.g., CIFAR-100-C vs. CIFAR-10-C), the target domain typically exhibits sparser class-wise coverage. Augmentation enriches the target sample diversity while preserving semantic consistency, thereby mitigating overfitting to spurious patterns and enhancing robustness under severe and complex distribution shifts. Nevertheless, even without FA, our method still significantly outperforms the baseline model, further confirming the critical role of our decentralized adaptation framework.

\begin{table*}[t]
\centering
\caption{Classification error rate ($\downarrow$) on ImageNet-C under \textbf{Mixed Distribution Shifts} using ViT-Base backbone.}
\label{tab:performance_evaluation3}
\renewcommand{\arraystretch}{0.9}
\footnotesize
\resizebox{\textwidth}{!}{
\begin{tabular}{lcccccccccccccccc}
\toprule
 \textbf{Methods} & \makecell{Gauss.} & \makecell{Shot} & \makecell{Impul.} & \makecell{Defoc.} & \makecell{Glass} & \makecell{Motion} & \makecell{Zoom} & \makecell{Snow} & \makecell{Frost} & \makecell{Fog} & \makecell{Brig.} & \makecell{Contr.} & \makecell{Elast.} & \makecell{Pixel} & \makecell{JPEG} & \cellcolor{lightblue}\textbf{Avg.} \\ 
\midrule
\textbf{IN-C (VitBase-LN)} & 65.8 & 67.3 & 65.3 & 68.8 & 74.4 & 64.3 & 66.6 & 56.8 & 45.2 & 48.6 & 29.2 & 81.8 & 57.1 & 60.8 & 50.2 & \cellcolor{lightblue}60.2 \\
TENT  & 60.6 & 60.4 & 59.6 & 63.6 & 67.8 & 57.1 & 61.2 & 55.0 & 48.8 & 47.4 & 28.6 & 66.7 & 53.9 & 50.4 & 44.4 & \cellcolor{lightblue}55.0 \\
ETA  & 59.3 & {57.8} & {57.9} & \underline{58.8} & 62.8 & \underline{52.5} & {58.2} & 51.0 & 46.4 & \underline{44.2} & 28.8 & 58.3 & \underline{51.1} & 46.9 & \underline{41.9} & \cellcolor{lightblue}\underline{51.7} \\
AdaContrast  & 64.8 & 63.4 & 63.3 & 72.8 & 76.6 & 73.7 & 74.6 & 67.7 & 48.0 & 89.6 & 30.2 & 93.2 & 60.8 & 57.3 & 46.3 & \cellcolor{lightblue}65.5 \\
CoTTA  & 89.4 & 92.0 & 88.9 & 93.6 & 92.6 & 90.6 & 86.5 & 94.9 & 88.2 & 86.6 & 75.8 & 96.5 & 85.7 & 93.5 & 84.6 & \cellcolor{lightblue}89.3 \\
SAR  & \underline{58.9} & \underline{57.6} & \underline{57.6} & 59.4 & 63.6 & 53.0 & 58.5 & 52.3 & 47.1 & 45.4 & \underline{28.3} & 61.6 & 51.4 & 47.4 & 42.0 & \cellcolor{lightblue}52.3 \\
RoTTA & 64.4 & 65.6 & 63.7 & 67.6 & 71.3 & 59.8 & 64.1 & 52.7 &\underline{43.5} & 48.6 & \textbf{27.9} & 78.5 & 54.3 & 60.4 & 50.1 & \cellcolor{lightblue}58.2 \\
RDumb  & 59.7 & 58.5 & 58.5 & 60.0 & 64.1 & 54.0 & 59.0 & 52.0 & 46.7 & 44.5 & 28.6 & 61.2 & 51.9 & 48.3 & 42.6 & \cellcolor{lightblue}52.6 \\
DeYO & 60.0 & 58.6 & 58.8 & \underline{58.8} & \underline{62.4} & 61.9 & \textbf{50.9} & \underline{46.7} & 51.9 & 45.2 & 29.7 & \underline{55.7} & 51.6 & \underline{45.8} & 42.8 & \cellcolor{lightblue}52.1 \\
\rowcolor{lightpink} \textbf{FreDA (ours)} & \textbf{55.9} & \textbf{53.7} & \textbf{55.0} & \textbf{58.0} & \textbf{57.9} & \textbf{50.9} & \underline{57.4} & \textbf{45.5} & \textbf{42.9} & \textbf{43.9} & 29.5 & \textbf{51.7} & \textbf{47.8} & \textbf{41.6} & \textbf{40.7} & \textbf{48.8} \\
\bottomrule
\end{tabular}}
\label{tab:vitresult}
\end{table*}

\begin{figure}[t]
\centering
\includegraphics[width=0.99\linewidth]{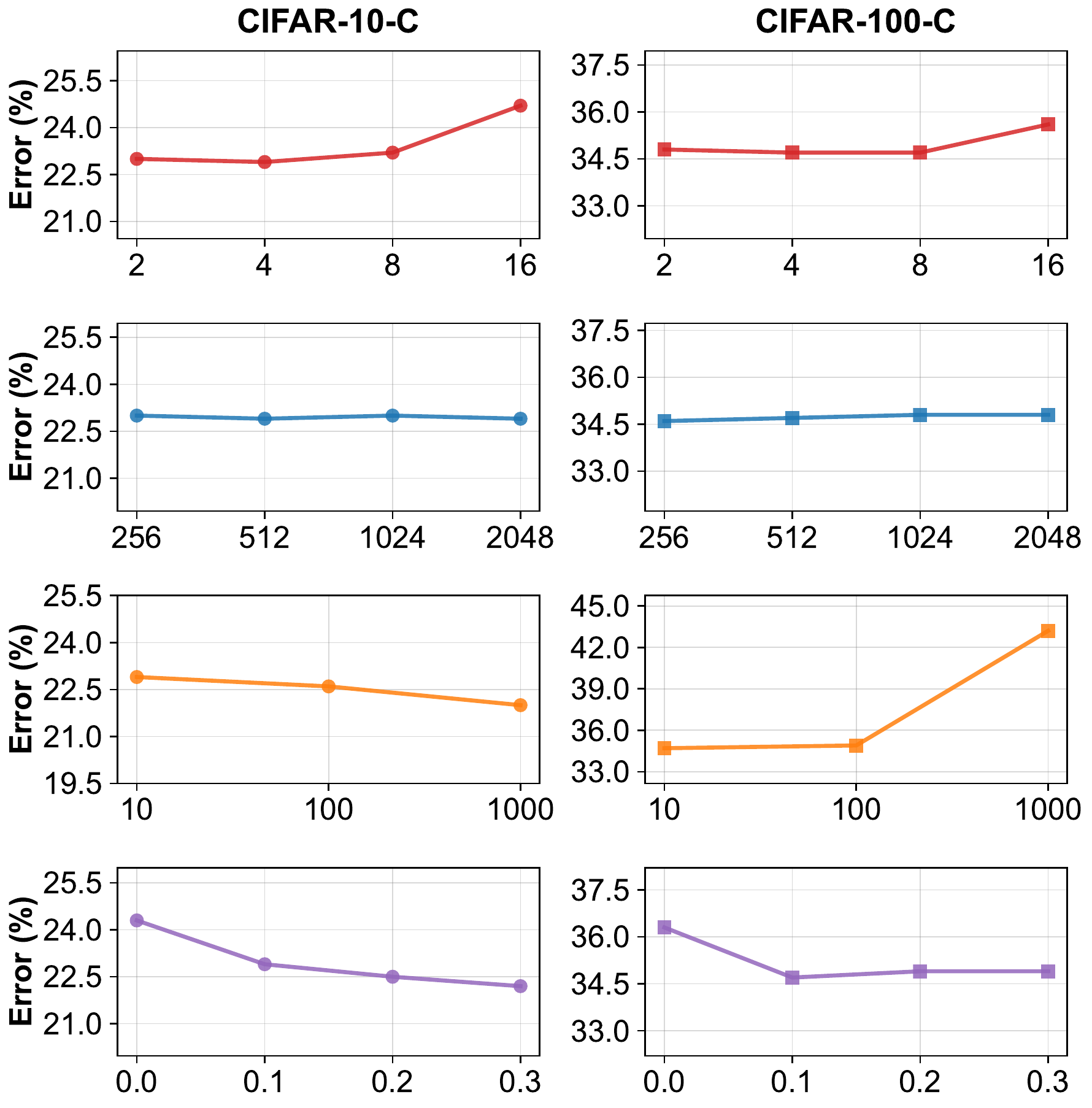}
\caption{Parameter study with respect to \texttt{CLUSTER\_NUM}, \texttt{KMEANS\_SIZE}, \texttt{COMM\_INTERVAL} \& \texttt{PERT\_MAGNITUDE} (from top to bottom) under \textbf{Mixed Distribution Shifts}.}
\label{fig:analyticalStudy}
\end{figure}

\subsubsection{FreDA adapts robustly to hyperparameter variations}
We investigate the impact of the key hyperparameters: \texttt{CLUSTER\_NUM}, \texttt{KMEANS\_SIZE}, \texttt{COMM\_INTERVAL}, and \texttt{PERT\_MAGNITUDE}. 
From Fig.~\ref{fig:analyticalStudy}, we have the following observations: 
\textbf{a)} \texttt{CLUSTER\_NUM} is largely insensitive in the range of 2–8: performance varies only marginally across CIFAR10/100-C and remains stable on average, with a clear drop emerging only at the extreme setting of 16. We attribute the degradation at 16 to over-partitioning, where the model tends to overfit to subtle, cluster-specific variations rather than capturing domain-invariant patterns. Therefore, a moderate number of clusters, around 4, appears to strike a good balance between adaptation flexibility and maintaining model robustness.
\textbf{b)} Varying \texttt{KMEANS\_SIZE} from 256 to 2048 results in stable performance across all datasets, indicating that our method is robust to changes in cluster sizes. 
\textbf{c)} Our approach shows general robustness to communication frequency (varying \texttt{COMM\_INTERVAL} from 10 to 1,000). On simpler datasets such as CIFAR10-C, infrequent communication (e.g., interval $f=1000$) performs best, likely due to effective independent learning. In contrast, for complex datasets (e.g. CIFAR100-C), more frequent communication (e.g., $f=10$) improves performance, likely by mitigating divergence among local branches and ensuring model consistency. 
\textbf{d)} Introducing perturbation (\texttt{PERT\_MAGNITUDE}) from 0 to a positive value yields a clear performance gain, confirming the effectiveness of our frequency-based augmentation strategy. Furthermore, when varying the magnitude between 0.1 and 0.3, FreDA maintains robust performance across datasets, indicating its resilience to moderate perturbation levels.

\begin{table}[t] 
\caption{Classification error rate ($\downarrow$) on CIFAR-10-C (C10), CIFAR-100-C (C100), and ImageNet-C (IN) using ResNet-50 \& ViT-Base backbones under \textbf{Continual Setting}, averaged over 15 corruptions.}
\centering
\footnotesize
\renewcommand{\arraystretch}{0.9}
\setlength\tabcolsep{8pt}
\resizebox{\linewidth}{!}{
\begin{tabular}{lcccc}
\toprule
\textbf{Methods} & \textbf{C10} & \textbf{C100} & \textbf{IN(ResNet)} & \textbf{IN(ViT)} \\
\midrule
Source          & 43.5      & 46.5       & {82.0}              & 60.2           \\
TBN           & 20.4     & 35.4      & 68.6             & -              \\
TENT  & 20.0     & 62.2      & 62.6             & 54.5           \\
ETA   & 17.9      & \underline{32.2}       & 60.2    & \underline{49.8}           \\
AdaContrast  & 18.5 & 33.5     & 65.5              & 57.0  \\
CoTTA  & \textbf{16.5}     & 32.8       & 63.1             & {77.0}\\
SAR    & 20.4     & \textbf{32.0}     & \underline{61.9}              & 51.7          \\
RoTTA  & 19.3     & 34.8     & 67.3            & 58.3          \\
RDumb  & \underline{17.8} & 34.1 &   {90.6}    & 50.2        \\
DeYO  & {87.0} & 98.1 & 90.6 & 94.3           \\
UnMix-TNS  & 24.9 & 32.7 & 75.4           & -              \\
\rowcolor{lightpink} \textbf{FreDA (ours)} & 19.5 & 32.5 & \textbf{60.2} & \textbf{47.9} \\
\bottomrule
\end{tabular}}
\end{table}

\subsection{Performance with Transformer Backbone}
In addition to evaluating our model on commonly compared CNN backbones, we further assess its performance under transformer-based architecture, specifically using the ViT-Base backbone. Here, we report results on the ImageNet-C benchmark using ViTBase-LN~\cite{dosovitskiy2020image} (see TABLE~\ref{tab:vitresult}), where the pretrained weights are obtained from \texttt{torchvision}. Importantly, all experimental configurations are kept consistent with those used in the CNN-based experiments. As shown, our method continues to deliver strong performance, demonstrating its robustness and adaptability across different backbone architectures.

\subsection{Performance under Continual Settings}
Although our method is specifically designed for mixed domain scenarios, we also evaluated its performance under the conventional continual test-time adaptation~\cite{wang2022continual,wang2021tent} setting to assess its robustness in different contexts. In this setting, the model adapts online to a sequence of test domains without explicit knowledge of domain shifts, with only one distribution shift occurring at a time and not reappearing. Without adjusting any parameters, our method demonstrated competitive performance compared to current state-of-the-art approaches. Notably, while UnMix-TNS effectively addresses non-i.i.d. issues (dependent sampling at the class level), it is less effective under i.i.d. conditions. Our results suggest that the proposed FreDA not only excels in its intended mixed domain scenarios but also generalizes effectively to standard continual adaptation tasks, providing a robust solution across various distributional challenges.

\section{More Details and Discussion}
\subsection{Adaptation Scenarios}\label{appdenix_adaptation_scenarios}
\subsubsection{Mixed Domains}
\noindent
{In this scenario, the model processes a long sequence of test samples where each sample $x_i \sim p_{e_i}(x)$ is independently drawn from a randomly selected target domain $\mathcal{D}_{e_i} \in \{\mathcal{D}_{t_1}, \mathcal{D}_{t_2}, \dots, \mathcal{D}_{t_N}\}$ and a randomly selected class $c_i$ among classes $\{1, 2, \dots, C\}$. The sequence is represented as:
\[
\left\{
x_1^{\mathcal{D}_{e_1},\, c_1},\ 
x_2^{\mathcal{D}_{e_2},\, c_2},\ 
\ \dots,\ 
x_k^{\mathcal{D}_{e_k},\, c_k}
\right\},
\]
where each target domain index $e_i \in \{1, 2, \dots, N\}$ and class number $c_i \in \{1, 2, \dots, C\}$ are independently and randomly selected for each sample $x_i$.}

\subsubsection{Continual Domain Adaptation}
In this setting, the model adapts online to a sequentially presented series of test domains, where each domain shift occurs only once and does not reappear. The sequence progresses through distinct target domains in a fixed order 
$D_1 \to D_2 \to \dots \to D_N$, with samples from each domain appearing contiguously. The class labels within each domain are independently and randomly selected. The sequence is structured as:

\begin{equation}
\begin{split}
&\underbrace{\{(x_1^{D_1, c_1}), (x_2^{D_1, c_2}), \dots, (x_{l_1}^{D_1, c_{l_1}})\}}_{\text{Samples from domain } D_1} \\&
\to
\underbrace{\{(x_{l_1+1}^{D_2, c_{l_1+1}}), \dots, (x_{l_1+l_2}^{D_2, c_{l_1+l_2}})\}}_{\text{Samples from domain } D_2} \\&\to
\dots \to
\underbrace{\{(x_{l_1+\dots+l_{N-1}+1}^{D_N, c_{l_1+\dots+l_{N-1}+1}}), \dots, (x_k^{D_N, c_k})\}}_{\text{Samples from domain } D_N}
\nonumber
\end{split}
\end{equation}

where:
\begin{itemize}
    \item $D_i$ denotes the $i$-th target domain in the fixed sequence, with $i \in \{1,2,\dots,N\}$.
    \item $c_m \in \{1,2,\dots,C\}$ is the randomly selected class label for the $m$-th sample, independent of domain transitions.
\end{itemize}

\subsection{Further Remarks on Related Works}

\subsubsection{Relation to Frequency Domain Learning}

Recent advances highlight frequency-based techniques as powerful tools for domain transfer. In domain generalization, frequency analysis has revealed critical insights into model robustness and learning dynamics~\cite{freq1,freq2,freq3,freq4,freq5,vaish2024fourier,freq6,freq7}. 
For domain adaptation, interpolating image amplitude spectra across styles has proven effective in reducing domain gaps and preventing overfitting to low-level statistics~\cite{fda1,fda2,fda3,fda4}. 
Motivated by these advancements, we discover that frequency information inherently captures domain characteristics, making it a valuable medium for decoupling mixed target domains -- an aspect largely unexplored in prior work. On this basis, we propose FreDA, addressing TTA under mixed domain shifts in Fourier space, with a decentralized adaptation and perturbation mechanism.

\subsubsection{Relation to Multi-Target Domain Adaptation}
TTA under mixed distribution shifts shares similarities with multi-target unsupervised domain adaptation (MT-UDA)~\cite{mt1,mt2,mt3,mt4,ocda1,ocda2}, yet diverges critically in complexity: while MT-UDA assumes static, predefined target domains and leverages labeled source data for explicit domain alignment, TTA operates with no access to source data and must adapt to dynamic, unpredictable target streams in an online manner. This eliminates direct source-target discrepancy computation and demand robust incremental adaptation rather than offline multi-domain optimization. These differences highlight the need for novel methodologies tailored specifically to TTA, going beyond the solutions developed for MT-UDA.

\begin{figure}[t]
  \centering
  \includegraphics[width=0.48\textwidth,clip, trim=5cm 4.5cm 7cm 4cm]{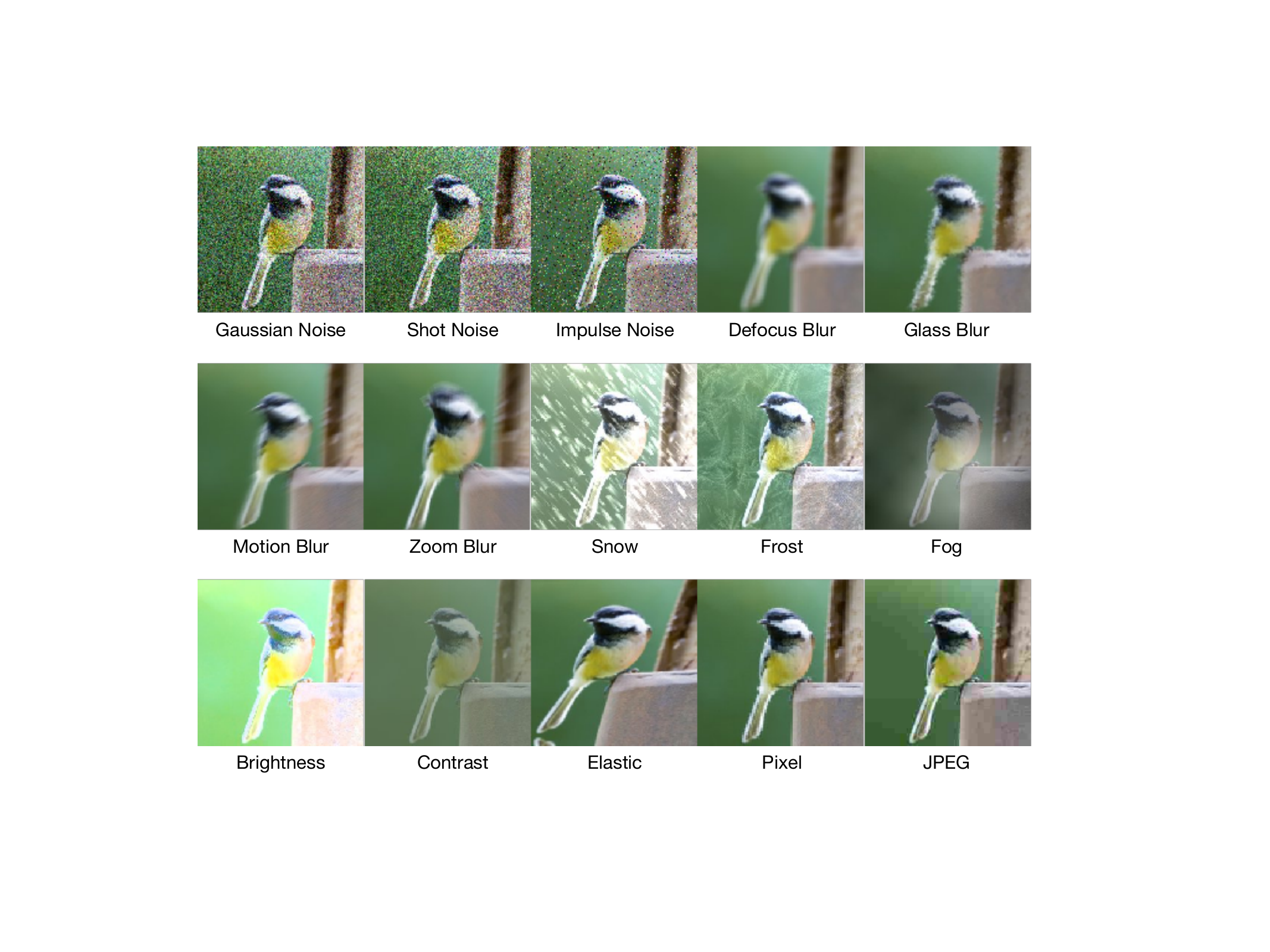}
  \caption{Examples from ImageNet-C under image corruptions. The images showcase a range of corruption types (e.g., noise, blur, and weather distortions) at varying severity levels.}
  \label{corruption}
\end{figure}

\begin{figure}[t]
  \centering
  \includegraphics[width=0.48\textwidth,clip, trim=4cm 3.8cm 9cm 3cm]{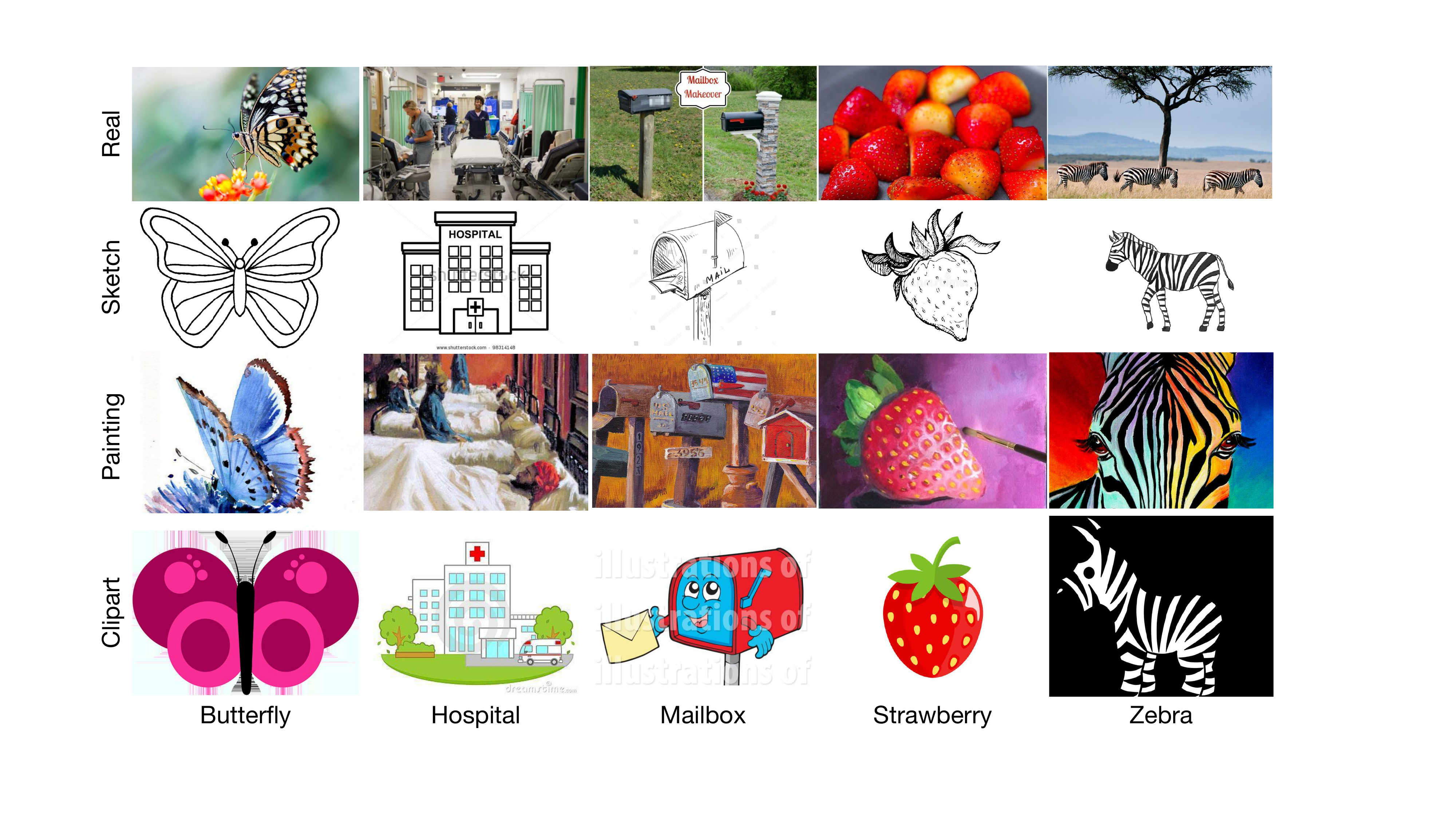}
  \caption{Samples from DomainNet126 across four subdomains (Real, Sketch, Painting, Clipart). These visualizations reflect the stylistic and perceptual variations inherent in each domain.}
  \label{domainnet}
\end{figure}

\subsubsection{Relation to Decentralized, Federated, Distributed Learning}
This work also intersects with decentralized, federated, and distributed learning due to splitting data batches into disjoint subsets and applying decentralized model adaptation.
First, while decentralized learning focuses on non-i.i.d. data that is naturally distributed across multiple nodes~\cite{hsieh2020non}, {our approach starts with a centralized batch of target samples. By proactively splitting this data into disjoint subsets, we expose latent non-i.i.d. characteristics, enabling the effective use of decentralized learning techniques.}
Second, federated learning considers data privacy and model collaborations within decentralized learning~\cite{fedavg}. In our case, as target samples are mixed in a batch, data privacy is not a concern. However, similar to federated learning, our approach also involves weight aggregation over subnetworks to enhance their base models.
Third, distributed learning aims to improve training efficiency on large-scale datasets by partitioning data for synchronized training~\cite{mcdonald2010distributed}. In contrast, our method operates in a real-time fine-tuning context with limited data at one time, hence scalability is less of a concern.

\subsection{Dataset Visualization}\label{apdix:data_visual}
To further illustrate the characteristics of the datasets used in our evaluation, we present visualizations of the data distribution across different corruption types (Fig.~\ref{corruption}), natural domain shifts (Fig.~\ref{domainnet}), and medical centers (Fig.~\ref{fig:high_freq_effect}). These figures highlight the diverse challenges that our models face in each evaluation scenario, providing insight into the complexity of the test conditions.

\section{Conclusion}
This paper advances Test-Time Adaptation by tackling the practical challenges posed by heterogeneous data streams. We introduce FreDA, a frequency-based decentralized adaptation framework that partitions incoming samples and performs localized adaptation, while employing {distribution-aware} augmentation to effectively handle mixed distribution shifts. Empirical results across multiple benchmarks demonstrate the effectiveness of our approach, highlighting its potential to drive future research on adapting to dynamic and diverse domain shifts.

\noindent \textbf{Future Work:}
FreDA provides a principled framework for adapting to mixed distribution shifts, and we believe several promising directions could further enhance its capabilities. 
{First, beyond the current proactive clustering, future work could investigate adaptive, learnable partitioning strategies that dynamically refine subset boundaries and better capture subtle relationships among target subdomains.}
Second, our frequency-based augmentation is currently applied at the sample level, which is well-suited for classification. Extending it to patch-level operations in spatial–frequency space may enrich the model’s granularity and open new possibilities for fine-grained tasks such as segmentation. Together, these directions offer natural extensions to broaden FreDA’s applicability in increasingly realistic and dynamic environments.
\bibliography{freda}
\bibliographystyle{IEEEtran}

\end{document}